%% file: main.tex
\newcommand\papername{\textsc{Com2Sense}}
\newcommand\numsamples{4k}

\newcommand\nuan[1]{{\color{olive}[{#1}]$_{nw}$}}

\input{macros.tex}

\documentclass[11pt,a4paper]{article}
\usepackage[hyperref]{acl2021}
\usepackage{times}
\usepackage{latexsym}
\usepackage{subfiles}
\usepackage{graphicx}
\usepackage{booktabs}
\usepackage{diagbox}
\usepackage{slashbox}
\usepackage{float}
\usepackage{textcomp}
\usepackage{enumitem}
\usepackage{xcolor} 
\usepackage{todonotes}
\usepackage{multirow}
\usepackage{subcaption}
\usepackage{hyperref}

\newcommand{\SideNote}[2]{\todo[color=#1,size=\small]{#2}} %

\newcommand{\telinwu}[1]{\SideNote{green!40}{#1 --Te-Lin}}

\newcommand{\yu}[1]{\SideNote{pink!40}{#1 --Hope}}

\restylefloat{table}

\definecolor{greenT}{RGB}{18, 99, 2}
\definecolor{redF}{RGB}{161, 11, 8}

\definecolor{blueDomain}{RGB}{14, 55, 237}
\definecolor{brownScenario}{RGB}{115, 92, 8}
\definecolor{violetNumeracy}{RGB}{105, 0, 150}

\usepackage{microtype}

\aclfinalcopy 
\def\aclpaperid{636} 

\title{
    \papername{}: A Commonsense Reasoning Benchmark with Complementary Sentences
}

{\centering
    \author{
        Shikhar Singh$^{*1}$,
        Nuan Wen$^{*1}$,
        Yu Hou$^1$,
        Pegah Alipoormolabashi$^2$,
        \\
        \textbf{
        Te-Lin Wu$^3$,
        Xuezhe Ma$^1$,
        Nanyun Peng$^3$
        }
        \\
    $^1$University of Southern California, $^2$ Sharif University of Technology,\\
    $^3$University of California, Los Angeles\\
    \texttt{\{ssingh43,nuanwen,houyu,xuezhe.ma\}@usc.edu}\ \
    \texttt{palipoor976@gmail.com} \\
    \texttt{\{telinwu,violetpeng\}@cs.ucla.edu} \\
    
}
}

\date{}

\begin{document}
\maketitle

\newcommand\blfootnote[1]{%
  \begingroup
  \renewcommand\thefootnote{}\footnote{#1}%
  \addtocounter{footnote}{-1}%
  \endgroup
}
\textbf{\blfootnote{$^*$ indicates equal contributions}}

\begin{abstract}
\input{0_abstract}
\end{abstract}

\input{1_intro}
\input{2_dataset}
\input{3_expt_setup}
\input{4_results_analysis}
\input{5_related_works}
\input{6_conclusion}

\input{7_ethics}

\bibliography{acl2021_new}
\bibliographystyle{acl_natbib}

\clearpage

\input{appendix}

\end{document}

%% file: macros.tex
\newcommand{\ie}{\textit{i}.\textit{e}.\ }
\newcommand{\eg}{\textit{e}.\textit{g}.\ }

\newcommand{\secref}[1]{Section \ref{#1}}
\newcommand{\figref}[1]{Figure \ref{#1}}
\newcommand{\tbref}[1]{Table \ref{#1}}

\newcommand{\dotieconcat}[2]{
  \text{\raisebox{.8ex}{$\smallfrown$}}%
}
\newcommand{\mypar}[1]{\noindent\textbf{#1}}

%% file: 0_abstract.tex
Commonsense reasoning is intuitive for humans but has been a long-term challenge for artificial intelligence (AI). 
Recent advancements in pretrained language models have shown promising results on several commonsense benchmark datasets. 
However, the reliability and comprehensiveness of these benchmarks towards assessing model's commonsense reasoning ability remains unclear.
%
%
To this end, we introduce a new commonsense reasoning benchmark dataset comprising natural language true/false statements, with each sample paired with its complementary counterpart, resulting in \numsamples{} sentence pairs. 
We propose a pairwise accuracy metric to reliably measure an agent’s ability to perform commonsense reasoning over a given situation.
The dataset is crowdsourced and enhanced with an adversarial model-in-the-loop setup to incentivize challenging samples. 
To facilitate a systematic analysis of commonsense capabilities, we design our dataset along the dimensions of knowledge domains, reasoning scenarios and numeracy.
Experimental results demonstrate that our strongest baseline (UnifiedQA-3B),
after fine-tuning, achieves \texttildelow71\% standard accuracy and \texttildelow51\% pairwise accuracy, well below 
human performance (\texttildelow95\% for both metrics). The dataset is available at \url{https://github.com/PlusLabNLP/Com2Sense}.

%% file: 1_intro.tex

\section{Introduction}



The capability of acquiring and reasoning over commonsense knowledge plays a crucial role for artificial intelligence (AI) systems that interact with humans and accomplish tasks in the real world.
For example, given a situation where \textit{someone is asleep}, an agent should choose to \textit{broom} instead of \textit{vacuum} to clean the room, as the latter would be noisy.
Likewise, a personal assistant should be able to infer that one is probably unavailable if they are \textit{at work}.
This ability to contextualize and draw upon implicit knowledge, and generalize to novel situations, requires commonsense reasoning.



\begin{figure}[t]
    \centering
    \includegraphics[width=\columnwidth]
    {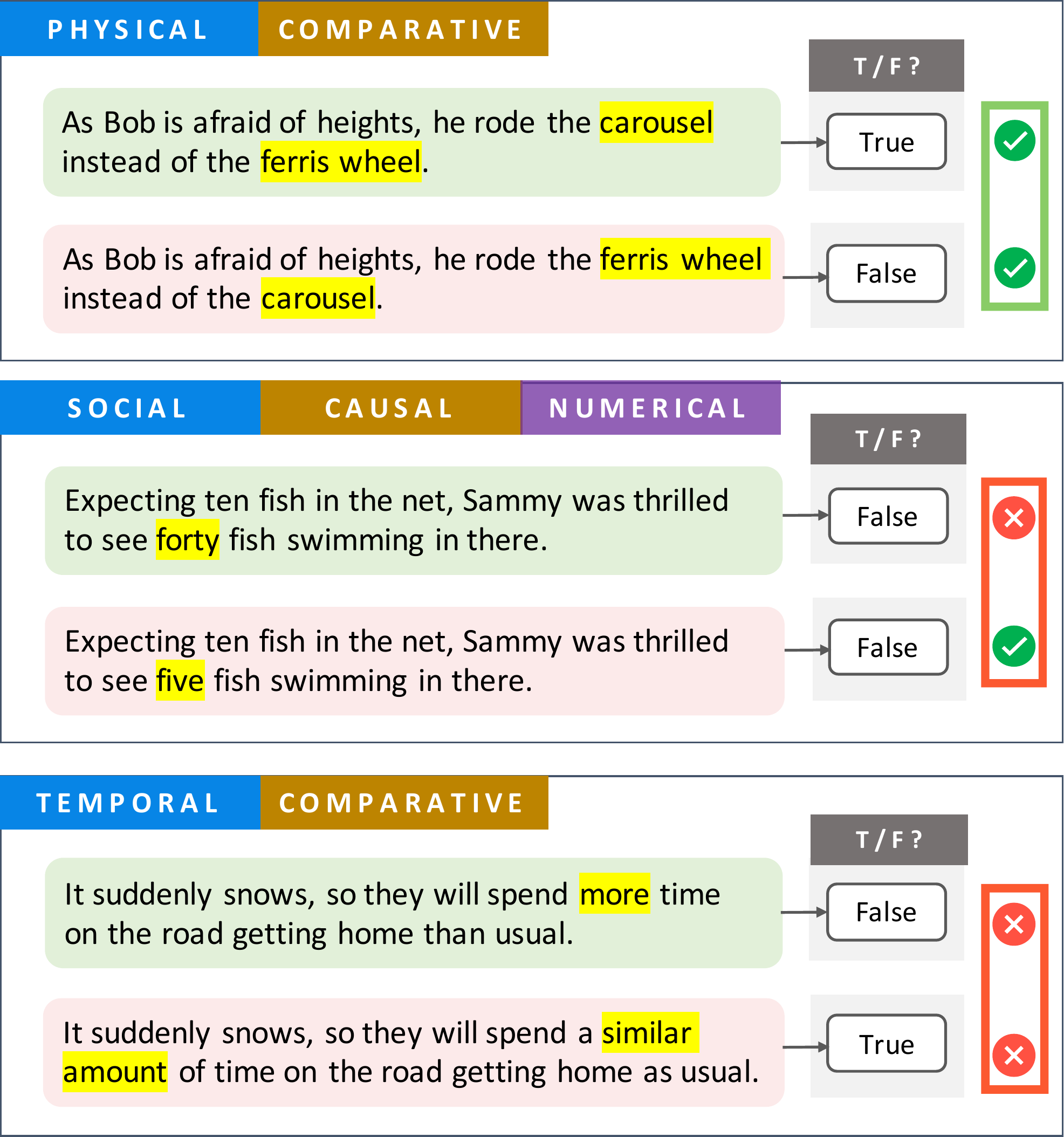}
    \caption{
    Complementary sentence pair samples from~\papername{} defined along \textcolor{blueDomain}{knowledge domains} (\eg \textit{physical}), \textcolor{brownScenario}{reasoning scenarios} (\eg \textit{comparative}) and \textcolor{violetNumeracy}{numeracy} attributes.
    Each sentence within a pair is either true (green boxes) or false (red boxes), followed by model predictions and annotations of whether the predictions are correct.
    A \textit{standard} accuracy is computed by the percentage of correctly judged sentences (50\% for these three pairs), while the \textit{pairwise} accuracy requires both \textit{individual} judgements to be correct in each pair (33\% for these three pairs).
    }
    \vspace{-1em}
    \label{fig:intro-fig}
\end{figure}



While humans are able to intuitively acquire commonsense knowledge from everyday experience and make sound inferences, whether current AI systems also possess such capabilities remains an open question. 
Recent advancements in natural language processing (NLP) has led to a surge in new benchmark datasets towards evaluating commonsense reasoning.
Specifically, existing benchmarks
are formulated as natural language inference 
(NLI)~\cite{bhagavatula2020abductive}, \textit{multiple choice} (MC) question answering~\cite{talmor-etal-2019-commonsenseqa, zellers-etal-2019-hellaswag, Bisk2020PIQARA}, and machine reading comprehension~\cite{huang2019cosmos} tasks.

While recent state-of-the-art models~\cite{liu2019roberta, raffel2020exploring, khashabi2020unifiedqa} have quantitatively demonstrated near human-level performance on these benchmarks, the exploitation of certain spurious patterns~\cite{gururangan-etal-2018-annotation, poliak-etal-2018-hypothesis, mccoy-etal-2019-right} in the datasets can be partly attributed to such achievements.
Consider the examples in~\figref{fig:intro-fig}, where each sentence is true/false, and is paired with a similar (with a few modifications) complementary counterpart such that the answer is flipped.
Humans can infer each statement independently with confidence, 
but models on the other hand struggle to give consistent judgements for the complementary pairs.
This indicates that models are able to \textit{guess} the correct answer without a thorough understanding of the given input.
If we formulate this as a multiple choice task, where \textit{only the true} sentence needs to be \textit{singled out} given the pairs, the models have higher chances to get it correct, as they are only required to select the~\textit{relatively better} option.

Furthermore, most existing commonsense benchmarks focus on the \textit{factual} aspects of commonsense~\cite{talmor-etal-2019-commonsenseqa, Bisk2020PIQARA}, and generally do not explicitly concern with \textit{reasoning} \cite{singer1992validation},
\ie the mental manipulation of factual knowledge, which we hypothesize is crucial for generalizing to novel situations.
While some prior works investigate commonsense reasoning in the context of social intelligence and co-reference resolutions~\cite{Sap2019SocialIC, Sakaguchi2020WINOGRANDEAA}, the reasoning components are implicit. 
Existing benchmarks fail to provide a systematic and comprehensive means of analyzing different aspects of commonsense knowledge and reasoning.
To address these challenges, we introduce the \textbf{Com}plementary \textbf{Com}mon\textbf{sense} (\papername{}) benchmark dataset which contains \numsamples{} complementary true/false sentence pairs. Each pair is constructed with minor perturbations to a sentence to derive its complement such that the corresponding label is inverted (see~\figref{fig:intro-fig}).
This \textit{pairwise} formulation provides a more reliable evaluation metric, where a model is considered correct \textit{only if it succeeds on both statements}.
We employ an adversarial crowdsourcing framework to collect human created samples via a \textit{gamified} machine-in-the-loop process:
A strong pretrained model is setup to provide instant feedbacks, thereby incentivizing challenging samples that can \textit{fool} the model. 
Broadly inspired by the \textit{Theory of Core Knowledge}, \ie the ability to reason about objects, places, numbers and the social world~\cite{spelke2007core}, we design our dataset along the following dimensions: \textbf{knowledge domains} (\textit{physical}, \textit{social}, \textit{temporal}), and \textbf{reasoning scenarios} (\textit{causal}, \textit{comparative}).
Additionally, concurrent to
a recent work~\cite{lin-etal-2020-birds} on studying numerical
commonsense, we include a third dimension of \textbf{numeracy}, which extends the factual focus of~\citet{lin-etal-2020-birds} (\eg \textit{``Ants have \textbf{six} legs.''}) to \textit{numerical reasoning} (\eg the \textit{ten fish} versus \textit{forty fish} in~\figref{fig:intro-fig}).
To the best of our knowledge, we are the first to explicitly introduce these dimensions in a commonsense benchmark dataset, thereby facilitating a more detailed and systematic probing of models' commonsense understanding.

Our experiments demonstrate that the best performing pretrained language models achieve \texttildelow71\% standard and  \texttildelow51\% pairwise accuracy, well below human performance. 
Additionally, we provide ablation studies on effect of training size on model performance, and the transferrability across the reasoning scenarios.
We summarize our contributions as follows:
1) We introduce a commonsense reasoning dataset
which we position as a challenging \textit{evaluation benchmark} (instead of a training resource) for NLP models.
2) We propose a pairwise evaluation metric featured by our complementary pair formulation for a more reliable assessment of commonsense reasoning abilities.
3) We benchmark state-of-the-art models that highlight significant gaps (\textgreater45\%) between model and human performances.

%% file: 2_dataset.tex
\section{Dataset}

\begingroup 
\renewcommand{\arraystretch}{1.3}
\begin{table*}[!h]
\centering
\small
\begin{tabular}{|c|c|c|l|c|}
\hline
 \textbf{Domain} & \textbf{Scenario} & \textbf{Numeracy} & \textbf{Example} & \textbf{Complement} \\ 
\hline

\multirow{2}{*}{Physical} & \multirow{2}{*}{Comparative} & \multirow{2}{*}{No} 

        & \emph{If we dropped \textbf{\textcolor{greenT}{milk}} on the floor, it is better to clean with} & \multirow{2}{*}{\textit{\textcolor{redF}{cereal}}} \\
&   &   & \emph{a mop rather than a broom.} &  \\
\hline


Physical & Causal & No & \emph{To \textbf{\textcolor{greenT}{read books}} at night, one should turn on the lights.} & \textit{\textcolor{redF}{see stars}} \\ 
\hline

\multirow{2}{*}{Social} & \multirow{2}{*}{Comparative} & \multirow{2}{*}{No}

        & \emph{Sam robbed a store, while Tim jumped the lights.}    & \multirow{2}{*}{\textit{\textcolor{redF}{chastising}}} \\ 
&   &   & \emph{People will likely be more \textbf{\textcolor{greenT}{forgiving}} towards Tim.}     &  \\
\hline

\multirow{2}{*}{Social} & \multirow{2}{*}{Causal} & \multirow{2}{*}{Yes} 

        & \emph{Given his \$1500 monthly income and no savings,}        & \multirow{2}{*}{\textit{\textcolor{redF}{\$3000}}} \\  
&   &   & \emph{he can afford an apartment rent of \textbf{\textcolor{greenT}{\$500}}.}      & \\
\hline

\multirow{2}{*}{Temporal} & \multirow{2}{*}{Comparative} & \multirow{2}{*}{Yes} 

        & \emph{Tim needs to return home in 2 hours, so he would}       & \multirow{2}{*}{\texttt{swap}} \\
&   &   & \emph{prefer to \textbf{\textcolor{greenT}{hit the gym}} rather than \textbf{\textcolor{redF}{go hiking}}.}    & \\
\hline

\multirow{2}{*}{Temporal} & \multirow{2}{*}{Causal} & \multirow{2}{*}{Yes} 

        & \emph{If Leo earns \$100 per day, then by working from}       & \multirow{2}{*}{\textit{\textcolor{redF}{Wednesday}}} \\
&   &   & \emph{Monday to \textbf{\textcolor{greenT}{Friday}} his weekly income will be \$500.}        & \\
\hline

\end{tabular}
    \caption{
        Data samples from different categories in \papername{}. Each example is labelled as \textcolor{greenT}{true}, while its complement (\textcolor{redF}{false}) is generated by substituting or swapping the words in \textbf{bold} (in green or red font).   
    }
\label{tab:data_samples}
\end{table*}
\endgroup





We introduce~\papername{}, a dataset for benchmarking commonsense reasoning ability of NLP models. 
We use crowdsourcing to collect the dataset and supplemented with an adversarial \textit{model-in-the-loop} approach.
The key features of our development process are:
1) qualification quiz to filter and familiarize workers, 
2) \textit{gamified} creation tasks, and 
3) quality check by experts.
The details of dataset formulation and collection procedure, along with statistics are provided in the following sections.

\begin{figure*}[t]
    \center
  \includegraphics[width=.95\textwidth]{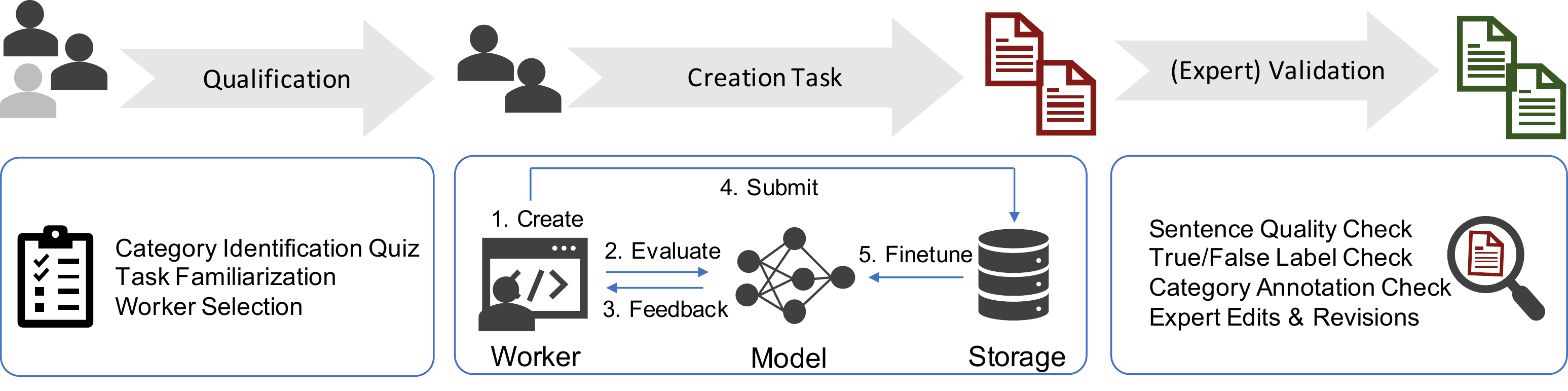}
  \caption{
    Data collection workflow: 
    1) qualification quiz to instruct the key aspects of our creation task and filter unqualified workers, 
    2) interactive \textit{model-in-the-loop} creation process to incentivize challenging samples via model feedback, and 
    3) data validation according to our guidelines and category descriptions.
  }~\label{fig:workflow}
  \vspace{-1em}
\end{figure*}

\subsection{Formulation}
\papername{} seeks to measure a comprehensive commonsense understanding of everyday events and entities. 
The task requires one to judge whether a given sentence is true or false. 
For each sentence in the dataset, we also compose its complementary counterpart by modifying a few words, such that the answer is inverted. 
The key advantages of using complementary pairs are two-folds: 1) it provides a more robust way of evaluating models' commonsense reasoning ability by requiring both sentences to be correctly judged, and 2) the complements naturally highlight the salient words which may be useful in probing model behaviors.

Furthermore, to facilitate a systematic study of commonsense, we design our dataset across the following three dimensions:

\begin{enumerate}[leftmargin=*]
\vspace{-0.5em}
    \item \textbf{Knowledge Domain:}
    We categorize commonsense knowledge into \textit{physical}, \textit{social} and \textit{temporal} domains. The physical domain emphasizes on an intuitive understanding of physical properties (\eg weight, shape, motion, space) and object affordances. The social domain encapsulates interactions (\eg intent, emotion, reaction), activities, and societal norms. The temporal domain captures the notion of time, particularly attributes such as duration, frequency and order of events. While domains may not always be strictly exclusive (\eg choice of transport and duration), our complementary pair setup naturally places emphasis on the intended domain.

    \item \textbf{Reasoning Scenario:}
    We define two types of inferential reasoning scenarios:
    1) The \textit{causal} scenario requires the ability to infer whether a cause explanation or a subsequent event (cause-effect) is correct. 
    2) The \textit{comparative} scenario requires the ability of determining the most plausible hypothesis between two or more competing ones. 

    \item \textbf{Numeracy:}
    Refers to the basic understanding of numbers, arithmetic, ratios, statistics, etc.
    With the objective of linking numeracy to commonsense, we particularly focus on ``number sense'' -- an intuitive understanding of numbers, their magnitude and relationships, rather than computational and numerical precision.
  
\end{enumerate}

\noindent Therefore, each sample in our dataset should fall into a category defined by a combination of the above dimensions, as exemplified in~\tbref{tab:data_samples}.


\subsection{Dataset Creation}

\papername{} is developed through crowdsourcing on Amazon Mechanical Turk (MTurk) with the goal of collecting complementary sentence pairs.
The creation tasks are constructed for each \textit{category} defined by the combination of domain, scenario and numeracy attributes.
An overview of the data collection workflow is illustrated in ~\figref{fig:workflow}.
In order to participate, the workers are required to pass a \textbf{qualification quiz} designed to familiarize them with the key aspects of our dataset.

\vspace{.3em}

\mypar{Creation:}
During the creation phase, to orient and aid workers' creativity, they are provided with five examples of complementary pairs that belong to a particular category as reference. We also share a list of verbs and topics pertinent to the current domain, as an optional resource. 
While our examples serve as a reference towards creating complementary pairs, the workers have the freedom to construct their sentences as they deem appropriate. 

We employ an adversarial \textit{model-in-the-loop} approach to provide workers with immediate feedback (\ie model predictions) on each created sentence.
After entering the inputs and labels, they may choose to evaluate and revise their inputs.
If the sentence successfully fools our model to answer incorrectly, workers are awarded with an additional amount for each input\footnote{Base pay = \$0.05 -- \$0.1 and bonus pay = \$0.5 -- \$0.9.}.

To further incentivize worker creativity, we offer bonuses if the inputs are qualitatively regarded as creative during the validation stage. 
Such \textit{gamified} process may continue for a few rounds until the workers are satisfied with their monetary rewards.

\vspace{.3em}

\mypar{Model:}
We deploy a RoBERTa-large based model for binary sentence classification, finetuned on SemEval-2020 Task 4 \cite{wang-etal-2020-semeval} given its true/false format and broad coverage of commonsense knowledge. 
After the first phase of collection (2k pairs), the model weights are updated by finetuning on our dataset with 60\% train, 20\% dev and 20\% test splits. 
This will naturally help diversify our dataset samples, as the model is unlikely to be fooled with repetitive knowledge and sentence structures.


\vspace{.3em}
\mypar{Validation:}
To ensure high quality, the samples are validated by internal members to look for inconsistencies with regard to the category-type \ie follows the domain, scenario and numeracy requirements, and inferential ambiguities that may arise due to insufficient context, specialized concepts, grammatical errors, etc. 
Furthermore, annotators may choose to revise the samples to fix any of the aforementioned issues.
Each sample is validated by three annotators and the final outcome is decided through a majority vote. 
The inter-annotator agreement score is 0.989 measured using Fleiss' Kappa.
Additionally, pairs in which neither input could fool the model are discarded during this stage.

\noindent The dataset is developed with the help of 173 workers. 
To ensure that workers are proficient in English, the demographic pool of the workers is initially limited to the United States. However, we removed this criteria to avoid cultural biases in the dataset.
Additionally, to understand the utility of our adversarial model feedback setting, we analyze the data on number of revisions made by workers in order to successfully fool the model.
We find that the average number of revisions is 1.36, while the median is zero.
This suggests that for majority of samples, workers find our reference material sufficient and are also able to leverage model feedback to aid their creations.
Additional details on dataset development are in Appendix~\secref{asec:dataset}.

\subsection{Dataset Statistics} 

Given that \papername{} is primarily a benchmark dataset, it is partitioned into train\footnote{As a resource to adapt models for our task.} (20\%), development (10\%), and test (70\%) set, respectively.
There are in total \numsamples{} of statement pairs in our dataset. Complementary statements from the same pair are distributed to the same partition.
\tbref{tab:dataset_stats} gives the essential statistics of our dataset across different splits. Note that due to the complementary pair formulation, the type-token ratio is approximately reduced by a factor of two, and the dataset is naturally balanced along the true and false labels.

\begin{figure}
    \centering
    \includegraphics[width=70mm,scale=0.5]{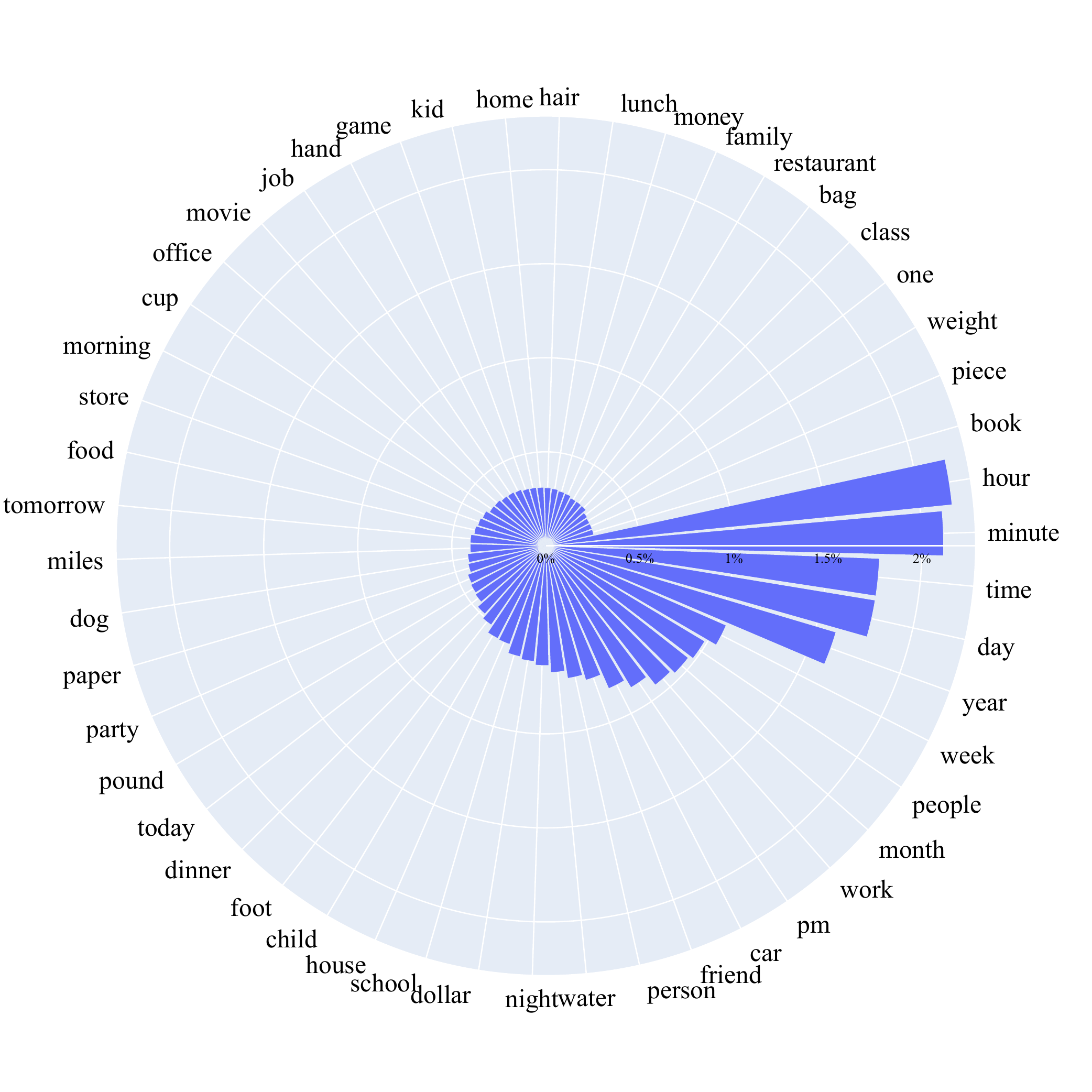}
    \vspace{-1em}
    \caption{
        Top-50 frequent nouns in the dataset.
    }
    \label{fig:radar-fig}
\end{figure}

\begin{figure}
    \centering
    \includegraphics[width=75mm,scale=0.9]{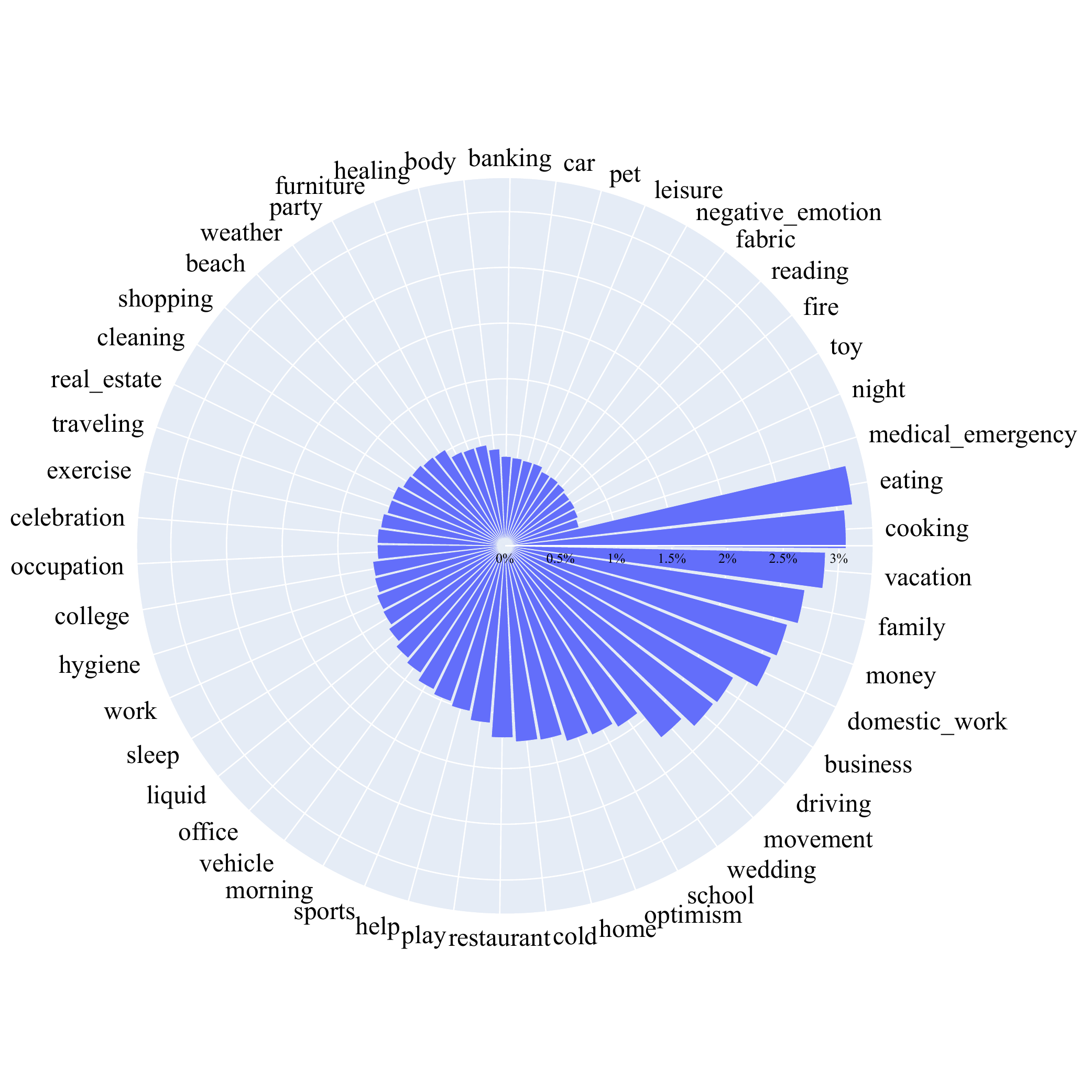}
    \vspace{-2em}
    \caption{
        Top-50 frequent topics in the dataset. 
    }
    \label{fig:topic_radar}
\end{figure}

\tbref{tab:dataset_category_wise} gives the breakdown of percentage of samples from each category defined by a combination of the three dimensions.
The distribution of most frequent nouns in the dataset is visualized in~\figref{fig:radar-fig}. Likewise, the distribution of most frequent topics -- lexical categories generated using the \textit{Empath}\protect\footnotemark tool, is provided in~\figref{fig:topic_radar}.

\footnotetext{https://github.com/Ejhfast/empath-client}

\begin{table}[h]
    \centering
    \small
        \begin{tabular}{@{}lrrrr@{}}
        \toprule
            {\bf Statistic}         & {\bf Train}   & {\bf Dev} & {\bf Test}    \\ \midrule
            \# complementary pairs  & 804           & 402       & 2779          \\
            Avg input length        & 21            & 21        & 21            \\
            Max input length        & 68            & 49        & 67            \\
            Min input length        & 6             & 7         & 6             \\
            \# unique tokens        & 2306          & 1541      & 4407          \\
            \# total tokens         & 21116         & 10520     & 72517         \\  \bottomrule
        \end{tabular}
    \caption{Dataset statistics across different splits\protect\footnotemark}.
    \label{tab:dataset_stats}    
\end{table}

\footnotetext{Input lengths are computed with Spacy tokenizer}

\begingroup
\setlength{\tabcolsep}{12pt} 
\begin{table}[h!]
	\centering
	\small
        \begin{tabular}{lcc}
            \toprule 
             & \multicolumn{2}{c}{\textbf{Scenario}} \\
                \cmidrule(l){2-3}
            \textbf{Domain}         & Causal            & Comparative  \\
            \midrule
                Physical            & 17.47\% (24\%)    & 18.92\% (23\%)    \\
                Social              & 14.68\% (50\%)    & 16.51\% (22\%)     \\
                Temporal            & 16.74\% (57\%)    & 15.68\% (62\%)     \\
            \bottomrule 
        \end{tabular}
	\caption{Category-wise breakdown (percentage) of dataset samples. The quantities in parenthesis refer to the relative proportion of samples with numeracy, under the given combination of domain and scenario.}
	\label{tab:dataset_category_wise}
\end{table}
\endgroup


%% file: 3_expt_setup.tex
\section{Experimental Setup}

\vspace{-.5em}
The experiments are designed to meet the following objectives:
1) benchmark state-of-the-art NLP models along the standard and pairwise formulation;
2) analyze the model performance across different categories of commonsense reasoning;
3) report the effect of training size on model performance;
and 4) verify the role of reasoning types by measuring ``cross-scenario" transferability.
\noindent Besides standard accuracy, we introduce a new metric called \textbf{pairwise accuracy} that evaluates as correct if both predictions within a pair are accurate. 

\noindent We benchmark several state-of-the-art NLP models, specifically the ones proven preeminent in existing commonsense benchmarks, and additionally include a Bi-LSTM model as a baseline to help check for potential spurious correlations in the dataset. 
\noindent We consider the following baselines:


\vspace{.3em}
\mypar{BiLSTM+GloVe}
A bidirectional-LSTM model~\cite{hochreiter1997long} taking input word embeddings from GloVe~\cite{pennington-etal-2014-glove}.

\vspace{.3em}
\mypar{BERT}
The BERT-base (110M) model introduced in~\cite{devlin-etal-2019-bert}.

\vspace{.3em}
\mypar{RoBERTa-large}
A large variant (355M) of RoBERTa model~\cite{liu2019roberta} built upon BERT-large architecture.

\vspace{.3em}
\mypar{DeBERTa-large}
Recently~\citet{he2020deberta} proposed a novel disentangled attention mechanism that improves upon BERT and RoBERTa models. 
We consider the large variant (390M) as a baseline. 

\vspace{.3em}
\mypar{T5-large}
Similarly, the large variant (770M) of the T5 model~\cite{raffel2020exploring}. 
We follow the standard prefix-based text-to-text format, and adapt it for our binary classification setup.

\vspace{.3em}
\mypar{UnifiedQA}
The UnifiedQA~\cite{khashabi2020unifiedqa} was originally trained on numerous datasets including several commonsense reasoning benchmarks, and performed well under zero-shot setting.
We consider the variants with T5-large and T5-3B as the architecture backbone.





%% file: 4_results_analysis.tex
\begin{figure*}[t!]
    \begin{subtable}{\columnwidth}
    \centering
      \includegraphics[width=\columnwidth]{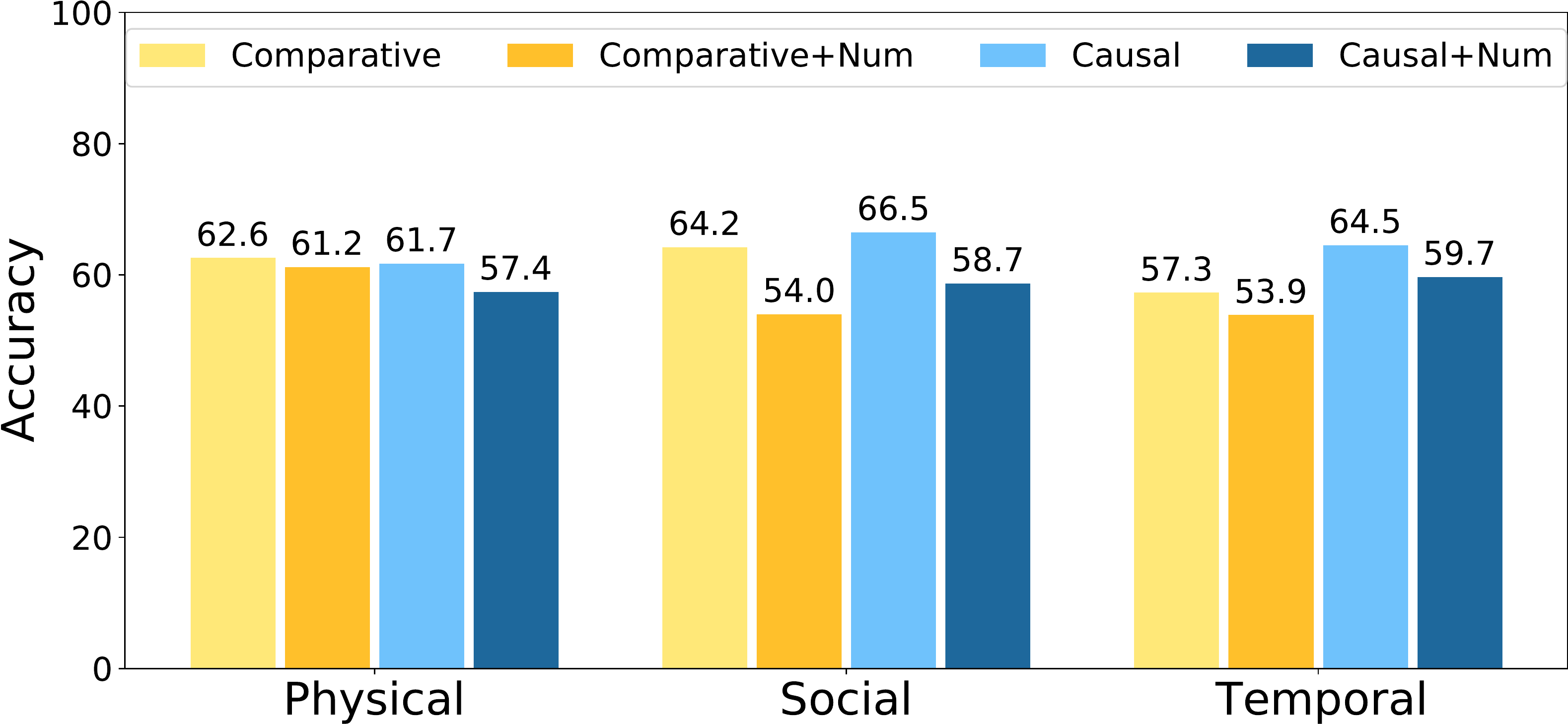}
        \caption{
            T5-large.
        }
        \label{fig:subsets_1}
    \end{subtable}%
\quad
    \begin{subtable}{\columnwidth}
    \centering
      \includegraphics[width=\columnwidth]{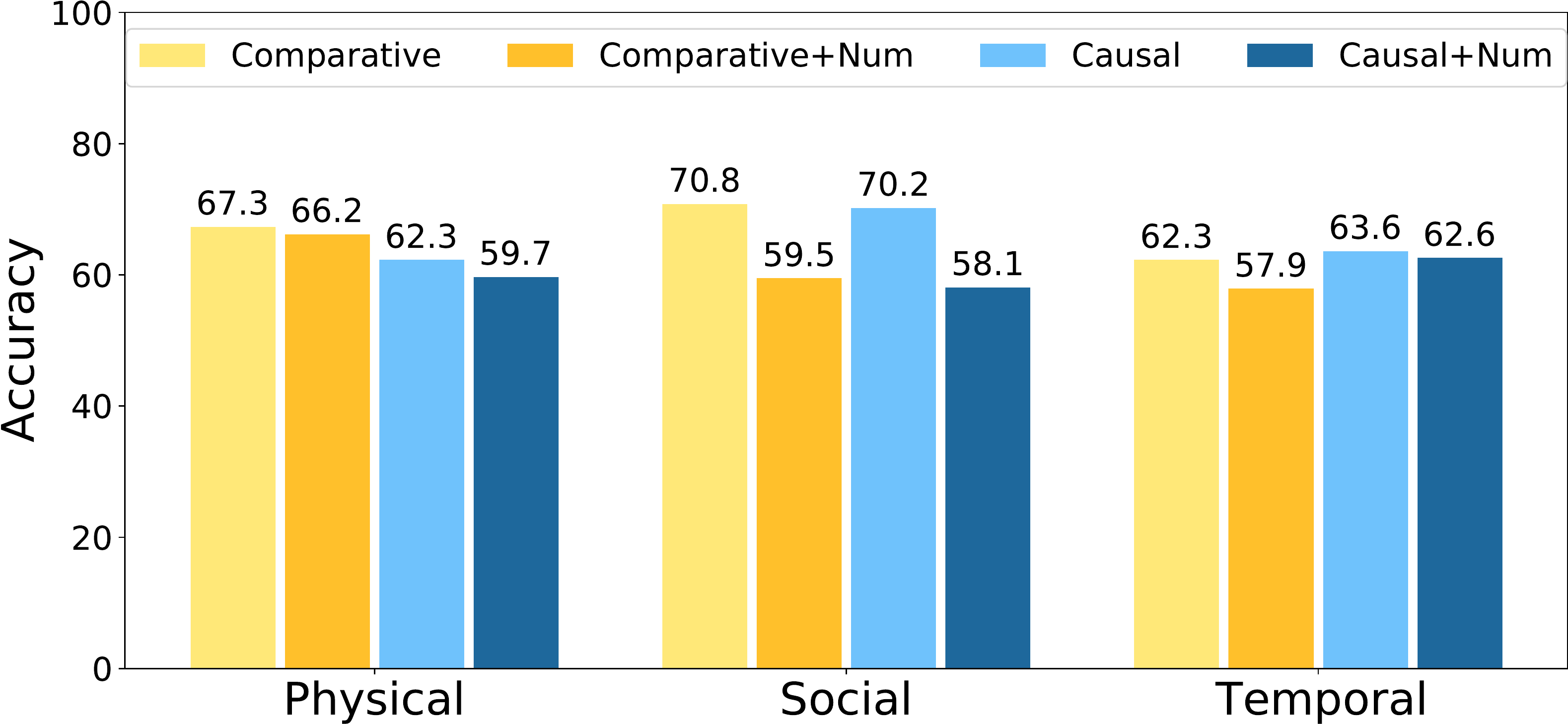}
        \caption{
            DeBERTa-large.
        }
        \label{fig:subsets_2}
    \end{subtable}
    
\caption{Model performance breakdowns across domains and scenarios, \texttt{"+Num"} denotes numeracy involved.}
\label{fig:subsets}
\end{figure*}

\begingroup
\setlength{\tabcolsep}{12pt} 

\begin{table}[t!]
	\centering
    \resizebox{0.9\columnwidth}{!}{
	\begin{tabular}{lcc}
		\toprule
		\textbf{Model}      & Standard  & Pairwise \\
		\midrule
		Random              & 50.00         & 25.00 \\
		BiLSTM+GloVe        & 53.80         & 29.50 \\
		BERT-base           & 51.79         & 12.91 \\   
		RoBERTa-large       & 59.35         & 33.28 \\
		T5-large            & 60.56         & 41.84 \\
		UnifiedQA-large     & 60.83         & 34.79 \\
		DeBERTa-large       & 63.53         & 45.30 \\
		UnifiedQA-3B        & \textbf{71.31} & \textbf{51.26} \\
		\cmidrule(lr){1-3}
		Human               & 96.50         & 95.00 \\
		\bottomrule
	\end{tabular}
	}
	    \caption{Test set accuracy for selected models, trained and evaluated on our dataset. Human performances are obtained with 200 randomly selected and decoupled samples from 100 pairs.}
	\label{tab:baseline_results}
\end{table}

\endgroup

\section{Results and Analysis}
\label{sec:main_results}

\mypar{Human Performance:}
To estimate a human upper bound for our dataset, we perform a separate run with ten top performing workers that had participated in our collection phase to examine a randomly selected subset of 200 samples (\ie from 100 pairs). 
Each worker is assigned with a set of shuffled samples with his/her own creations deliberately filtered out. The answer is determined by majority vote from \textit{three} workers. Human performances are 96.5\% with standard accuracy and 95.0\% with pairwise accuracy, respectively.




\subsection{Results}

\mypar{Benchmark results:} \tbref{tab:baseline_results} summarizes the baseline performances on the test set.
As the Bi-LSTM model performs close to random, we claim that improvements from stronger baselines should be attributed to the models and not annotation biases that they can exploit. Among the  baseline models, the UnifiedQA-3B achieves the best performance on both the standard and pairwise metric.
Note that the number of learnable parameters in UnifiedQA-3B is much larger than those in the second and third best models, which are DeBERTa-large (390M) and UnifiedQA-large (770M).
Our~\papername{} benchmark remains quite challenging, as there are significant gaps between the model and the human performances.


\begin{table}[t!]
	\centering
	\resizebox{\columnwidth}{!}{
	\begin{tabular}{lcccc}
		\toprule
		& \multicolumn{3}{c}{\textbf{Model}} \\ \cmidrule(lr){2-4}

		\textbf{Dataset} & Random  & RoBERTa & T5 & Human \\
		\midrule
		CQA                 & 20.00 & 72.10 & 73.35 & 88.90 \\
		SWAG                & 25.00 & 89.92 & 88.72 & 88.00 \\
		SocialIQA           & 33.33 & 77.12 & 73.25 & 84.40 \\
		PIQA                & 50.00 & 77.21 & 79.89 & 94.90 \\
		WinoGrande          & 50.00 & 79.14 & 75.02 & 94.00 \\
		\cmidrule(lr){1-5}
		\papername\textsubscript{standard}          
		                    & 50.00 & 59.35 & 60.56 & 96.50 \\
		\papername\textsubscript{pairwise}          
		                    & 25.00 & 33.28 & 41.84 & 95.00 \\
		\bottomrule
	\end{tabular}
	}
	\caption{
    	Test set accuracy for selected models (RoBERTa-large and T5-large), trained and evaluated on respective datasets.
        }
	\label{tab:dataset_comparison_results}
\end{table}

\vspace{.3em}
\mypar{Dataset comparisons:}
In order to contrast the difficulty of~\papername{} with other related benchmarks, we report the performances of two well performing models on the following:
CommonsenseQA (CQA) \cite{talmor2019commonsenseqa}, 
SWAG \cite{zellers-etal-2018-swag},
SocialIQA \cite{Sap2019SocialIC},
PhysicalIQA (PIQA) \cite{Bisk2020PIQARA} and
WinoGrande \cite{Sakaguchi2020WINOGRANDEAA}.

\noindent The results in~\tbref{tab:dataset_comparison_results} indicate that models clearly struggle more to perform well on ~\papername{} than other datasets.




\vspace{.3em}
\mypar{Performance across domains and scenarios:}
In~\figref{fig:subsets} we present the in-depth breakdown results for T5-large and DeBERTa-large across combinations of domain, scenario and numeracy. 
We observe that models consistently perform worse in categories involving numeracy, highlighting the limitations of current language models.
For physical domain, both models perform worse in causal than in comparative scenario.
We hypothesize that while the models may possess the required physical knowledge, they fail to generalize to a logical reasoning over known facts or grasp the implicit changes of physical properties, which is generally unseen in the pretraining corpora for NLP models.
Opposite trends are observed in both social and temporal domains, where similar hypothesis can be made that causal statements are more frequent patterns in the corpora when social activities or senses of time are the subjects.
Furthermore since model feedback was part of our dataset construction, we also report the category-wise difficulty in fooling the model (number of trials) during sample creation in~\secref{asec:adv} for a reference.


\begin{table}[t!]
    \centering
    \resizebox{0.65\columnwidth}{!}{
        \begin{tabular}{ccc} 
        \toprule
        \addlinespace
                \textbf{Setting}        & Test       & Dev      \\ 
        \midrule
                Multiple-Choice         & 70.63      & 77.61    \\
                Standard (T/F)          & 63.53      & 66.29    \\
        \bottomrule
        \end{tabular}
    }
    \caption{Performance (accuracy) of DeBERTa-large on the MC formulation of our dataset compared to the standard true/false setting. The model performs relatively worse on the latter.}
    \label{tab:mcq_results}
\end{table}

\vspace{.3em}
\mypar{True/False versus multiple choice setup:}
We further conduct an experiment with DeBERTa-large model on the same data splits with the input formulated as an \textbf{MC task} in~\tbref{tab:mcq_results}. 
Under this setting, the model is provided with the sentence pair and is required to select the true sentence among the two choices, for the response to be correct.
The performance is significantly higher compared to both standard  (\textgreater7\%) and pairwise accuracy (\textgreater25\%) presented in~\tbref{tab:baseline_results}. This result supports our intuition that it is easier for models to exploit spurious correlations in the surface patterns under the \textit{multiple choice} question answering setup.


\begin{table}[t!]
	\centering
	\resizebox{\columnwidth}{!}{
        \begin{tabular}{lcccc}
            \toprule 
            & \multicolumn{4}{c}{\textbf{Model\textsubscript{metric}}} \\
            \cmidrule(l){2-5}
            \textbf{Train set}      & DeBERTa\textsubscript{std}    & T5\textsubscript{std}        
                                    & DeBERTa\textsubscript{pair}   & T5\textsubscript{pair}   \\
            \midrule
                20\%                & 63.92     & 60.65     & 41.51     & 34.29     \\
                40\%                & 67.74     & 62.60    & 48.04      & 38.11    \\
                60\%                & 68.46     & 63.96     & 48.47     & 40.66     \\
            \bottomrule
        \end{tabular}
	}
	\caption{
        	Performance across different training set sizes (20\% / 40\% / 60\% of the entire dataset) for DeBERTa-large and T5-large. \texttt{"std"} and \texttt{"pair"} stand for standard and pairwise accuracy correspondingly.
        }  
	\label{tab:train_set_size_results}
\end{table}


\subsection{Analysis}

\mypar{The Effect of Training Data Size:} 
To study the effect of training size on model performance, we design an experiment by varying the sample size in the training set, with fixed dev (10\%) and test (30\%) sets to \textbf{ensure consistency} in evaluation.  
We consider DeBERTa-large and T5-large models for this ablation study, and report our findings in \tbref{tab:train_set_size_results}.
The results indicate a plateau in performance with increase in training samples.

\begin{table}[t!]
    \centering
	\resizebox{0.7\columnwidth}{!}{
    \begin{tabular}{ccc}
        \toprule
        \addlinespace
            \textbf{Setting}                & Standard  & Pairwise \\ 
        \midrule
            Train-C\textsubscript{exclude}  & 56.52     & 19.00  \\
            Train-C\textsubscript{include}  & 63.54     & 40.49  \\
        \bottomrule
\end{tabular}
}
\caption{Test set performance of DeBERTa-large on two different setups with respect to the complementary pairs. Both setups have the same training set size, but in Train-C\textsubscript{exclude} only one sentence of each pair is present in the training set, while in Train-C\textsubscript{include} both samples in a pair are included. The results indicate the effectiveness of training the models with our formulated complementary pairs. 
}
\label{tab:intrapair}
\end{table}

\vspace{0.3em}
\mypar{Role of Complementary Pairs:} 
In previous experiments, we measure the model generalizations by distributing data samples into train and evaluation sets \textbf{by complementary pair}.
This ensures the similarly constructed sentences within the same pair is not \textit{leaked} into different data splits, and thus an ``inter-pair'' generalization is measured.
To investigate the effectiveness of our complementary pair formulation on training models to acquire commonsense reasoning ability, 
we first sample a subset with identical size (20\% data, 800 pairs) to that of the original train set, and then construct the following two variants (using the same subset): 
\vspace{-0.5em}
\begin{itemize}[leftmargin=*]
    \itemsep0em
    \item \textbf{Train-C\textsubscript{exclude}:} One of the complementary samples (in each pair) is \textit{excluded}, \ie no two samples belong to the same pair in this train set. It comprises 800 samples from 800 pairs, with balanced true/false labels.
    \item \textbf{Train-C\textsubscript{include}:} We retain half of the data samples where both sentences from a pair are \textit{included}, which leads to 800 samples from 400 pairs in this train set.
\end{itemize}
\vspace{-0.5em}

\noindent The remaining samples from the dataset (without the excluded ones in the two settings) are then split into dev (10\%) and test (70\%) sets.
We compare the performance of DeBERTa-large in the above two different settings in~\tbref{tab:intrapair}.
The results show a significant decrease in performance when complementary sentences are not provided. We hypothesize that the worse performances are due to the models' tendency to pick up surface patterns and memorize the labels in the training set without really understanding the scenario. Also, model generalization benefits from having complementary samples within the training set.

\vspace{0.3em}
\mypar{Cross-Scenario Generalizability:}
Given that knowledge domain and numeracy attributes of our dataset are intuitively distinct, we intend to quantitatively investigate if the same holds for reasoning scenarios.
Our ``cross-scenario'' experiments with DeBERTa-large, \ie trained on \textit{causal}, evaluated on \textit{comparative} and vice versa, indicate a poor generalization across both standard and pairwise accuracy metrics (see~\tbref{tab:cross_scenario_results}), underscoring the significance of having reasoning types.

\begin{table}[t]
	\centering
	\resizebox{\columnwidth}{!}{
        \begin{tabular}{lcccc}
            \toprule 
            & \multicolumn{4}{c}{\textbf{Evaluate}} \\
            \cmidrule(l){2-5}
            \textbf{Train}      & Causal\textsubscript{std}     & Comp.\textsubscript{std}          
                                & Causal\textsubscript{pair}    & Comp.\textsubscript{pair} \\
            \midrule
                Causal          & \textbf{63.64}                & 59.36         
                                & \textbf{35.46}                & 28.25  \\
                                
                Comp.           & 58.47                         & \textbf{64.50}         
                                & 26.43                         & \textbf{43.86}  \\
            \bottomrule
        \end{tabular}
	}
	    \caption{
	        Performance of DeBERTa-large trained on X and evaluated on Y, where X and Y are partitions created as per a reasoning scenario \textit{(causal, comparative)}.
	    }  
	\label{tab:cross_scenario_results}
\end{table}


%% file: 5_related_works.tex
\section{Related Works}




\mypar{Commonsense Resources:}
As commonsense is a crucial component to the actualization of AI, there has been a surge in creating relevant benchmarks, notable ones include evaluating machines' commonsense abilities in the format of
pronoun resolution~\cite{Levesque2011TheWS, Sakaguchi2020WINOGRANDEAA},
multiple choice~\cite{zellers-etal-2018-swag, talmor-etal-2019-commonsenseqa},
natural language generations~\cite{lin-etal-2020-commongen},
story understanding~\cite{mostafazadeh2016corpus},
and reading comprehensions~\cite{zhang2018record, huang2019cosmos, ning2020torque}.
Our work puts forth to create a commonsense benchmark in the format of true/false complementary pairs, where a more robust pairwise accuracy is adopted.
Note that although natural language inference (NLI) can be tasked similarly to the true/false formulation, the existing commonsense NLI benchmark either is not crowdsourced with high quality~\cite{zhang-etal-2017-ordinal}, or still resorts to a multiple choice setting~\cite{bhagavatula2020abductive}.
There are also benchmarks that specifically concern a type of commonsense knowledge, such as physical~\cite{Bisk2020PIQARA} and social~\cite{Sap2019SocialIC} intelligence, as well as temporal understanding~\cite{zhou-etal-2019-going}.
The ability to understand and induce numerical knowledge in texts has been studied in several recent works~\cite{dua-etal-2019-drop, ravichander-etal-2019-equate}, including numerical commonsense~\cite{lin-etal-2020-birds}. Our work differs to these works in the focus on less factual and arithmetic-precise numerical knowledge, but more on the intuitive sense of numbers, in conjunction with our defined knowledge domains and the scenarios.

It is worth noting that some prior works~\cite{wu2017visual, clark2019boolq} also investigate the effectiveness in the binary true/false (yes/no) formulation to construct a question answering dataset, while~\papername{} is the first to focus on commonsense reasoning.

\vspace{.3em}
\mypar{Dataset Biases:}
It is a widely perceived issue that spurious statistical patterns in datasets can often be exploited by machine learning models, which can potentially lead to overoptimistic judgements on the model improvements.
Particularly in NLP domain, prior works have shown that hypothesis-only baselines or syntactic heuristics perform surprisingly well in the NLI task~\cite{gururangan-etal-2018-annotation, glockner2018breaking, tsuchiya2018performance, poliak-etal-2018-hypothesis, mccoy-etal-2019-right}. Model exploiting biases or failing on simple adversarial patterns, can also be seen in sentence classification~\cite{wieting2019no} and question answering~\cite{jia2017adversarial, kaushik-lipton-2018-much, GevaEtAl2019} tasks.
We put forth to reduce the potential sentence-level biases by requiring the models to perform equally well on both directions in a complementary true/false pair.

\vspace{.3em}
\mypar{Adversarial Data Collection:}
Removing representation biases in a dataset by adversarially filtering undesired data samples, has been frequently practiced to collect datasets more challenging  to the models.
Recent work \textit{AFLite}~\cite{Sakaguchi2020WINOGRANDEAA, le2020adversarial}, built upon the adversarial filtering (AF) method in~\cite{zellers-etal-2018-swag, zellers-etal-2019-hellaswag}, adopted an iteratively improving model-in-the-loop approach to collect challenging commonsense benchmarks~\cite{Sakaguchi2020WINOGRANDEAA, Bisk2020PIQARA}. \textit{Gamified}~\cite{yang2017mastering} or interactive~\cite{Wallace2019Trick} approaches leverage human-in-the-loop to increase the difficulty of datasets and hence more robust model training. Counterfactual editing of data samples with human annotators~\cite{kaushik2019learning, gardner2020evaluating} is also closely related to our complementary pair construction that seeks to \textit{invert} the model predictions for a more reliable evaluation. 

Recently, several works have attempted to exploit the merits in involving both models and humans in the data creation cycle, \ie human-and-model-in-the-loop, to construct data samples that are both \textit{new} and challenging to the models~\cite{chen-etal-2019-codah, nie-etal-2020-adversarial, bartolo-etal-2020-beat}.
To our best knowledge, we are the first to employ such an approach in constructing commonsense reasoning benchmark, specifically, our complementary pair formulation makes it more sophisticated as the annotators are required to not only \textit{fool} the model but also pay attention to the salient concepts of their creations in both directions.




%% file: 6_conclusion.tex
\section{Conclusion}


We present a new challenging commonsense reasoning benchmark,~\papername{}, developed via an adversarial \textit{gamified} model-in-the-loop approach.
\papername{} comprises \numsamples{} \textit{manually created} complementary true/false statement pairs, designed along three dimensions: knowledge domain, reasoning scenario, and numeracy. 
We propose a robust pairwise metric to evaluate models' commonsense reasoning ability based on the complementary pair formulation, and benchmark the dataset with several state-of-the-art NLP models, highlighting significant gaps well below human performances (\textgreater\ 45\% gap).

On top of providing a new commonsense reasoning benchmark, we demonstrate studies on transferrability among defined commonsense aspects, with an objective to spur future research on a more systematic probing of models' grasp of commonsense.
As a potential future work drawn from these insights, we hope to inspire future model developments, specifically in two directions: 1) the ability to reason over known facts (\ie reasoning scenario), and 2) acquiring the implicit knowledge that is commonsensible to humans (\ie knowledge domain).
Furthermore, we hope our investigation in the formulations of question answering task (\ie MC setting versus our true/false complementary setting) can shed light on future explorations in identifying potential artifacts in NLP datasets.

%% file: 7_ethics.tex
\section{Acknowledgements}

We thank the anonymous reviewers for their feedback, and all the workers who have participated in the dataset creation on Amazon Mechanical Turk.
We give special thanks to Peifeng Wang, Xiangci Li, and Gleb Satyukov for their great help in the early stage of annotation pipeline construction as well as providing valuable feedback in composing the guidline instructions.
This work is supported by the Machine Common Sense (MCS) program under Cooperative Agreement N66001-19-2-4032 with the US Defense Advanced Research Projects Agency (DARPA).
The views and the conclusions of this paper are those of the authors and do not reflect the
official policy or position of DARPA.

\section{Ethics and Broader Impacts}

We hereby acknowledge that all of the co-authors of this work are aware of the provided \textit{ACM Code of Ethics} and honor the code of conduct.
This work is mainly about the creation of a challenging commonsense benchmark dataset.
The followings give the aspects of both our ethical considerations and our potential impacts to the community.

\paragraph{Dataset} We collect an English dataset of commonsense complementary sentence pairs via Amazon Mechanical Turk (MTurk) and ensure that all the personal information of the workers involved (e.g., usernames, emails, urls, demographic information, etc.) is discarded in our dataset.
This research has been reviewed by the \textbf{IRB board} and granted the status of an \textbf{IRB exempt}.
The detailed annotation process (pay per amount of work, guidelines) is included in the appendix; and overall, we ensure our pay per task is above the the annotator's local minimum wage (\texttildelow \$12 USD/HR).
Although commonsense can vary from different demographic areas, we primarily consider English speaking regions for the first round, and include more annotators from non English-spoken countries to diversify the dataset.
Future work can include collecting a more diverse dataset across more demographics regions to incorporate more regional-dependent commonsense, while using some post editing to ensure English proficiency of the constructed data.

\paragraph{Techniques} We benchmark the created dataset with the state-of-the-art large-scale pretrained language models, with minimum adaptation to the formulation of this dataset (\ie true/false formulation).
As commonsense is of our main focus, we do not anticipate production of harmful outputs, especially towards vulnerable populations, after training NLP models on our dataset.

%% file: appendix.tex
\appendix


\section{Additional Details of~\papername{}}
\label{asec:dataset}

\subsection{Collection with MTurk}

\mypar{Qualification Quiz}
To familiarize the workers with our collection task, we design a quiz with the following types of questions:
1) examine if a given statement can be correctly judged with only commonsense or it requires specialized knowledge,
2) infer the true/false label of a given statement,
and 3) select the most suitable domain and scenario where a given statement belongs to.

\vspace{.3em}
\mypar{Human Intelligence Tasks (HITs)}
The general instructions of our HIT page include: \textbf{Task Overview}, \textbf{Task Payment} and \textbf{Overall Task Procedure} for each category, to engage more workers.
At the end of the HIT instruction page, a link is provided to redirect the workers to the data creation page, where more detailed instructions and useful resources for the creation tasks are provided.
Besides passing our qualification quiz, the workers are also required to have a \textit{HIT Approval Rate} greater than 98\% and the \textit{Number of HITs Approved} greater than 5000.
In each HIT assignment, workers are required to submit three complementary pairs.
In the first phase of data collection, the base pay is \$0.6 for each assignment and workers will receive a \$0.5 bonus per sentence if it follows our instructions and fools the model; for the second phase, the base payment for each assignment is \$0.3 but we change the bonus to: \$0.5 (for either high-quality sentence or successful fooling) or \$0.9 (if both requirements are met, similar to those for the \$0.5 bonus in the first phase) to encourage workers to create higher quality data.

\subsection{Details of the Creation}

\begin{figure*}
    \centering
    \includegraphics[width=160mm, scale=1]{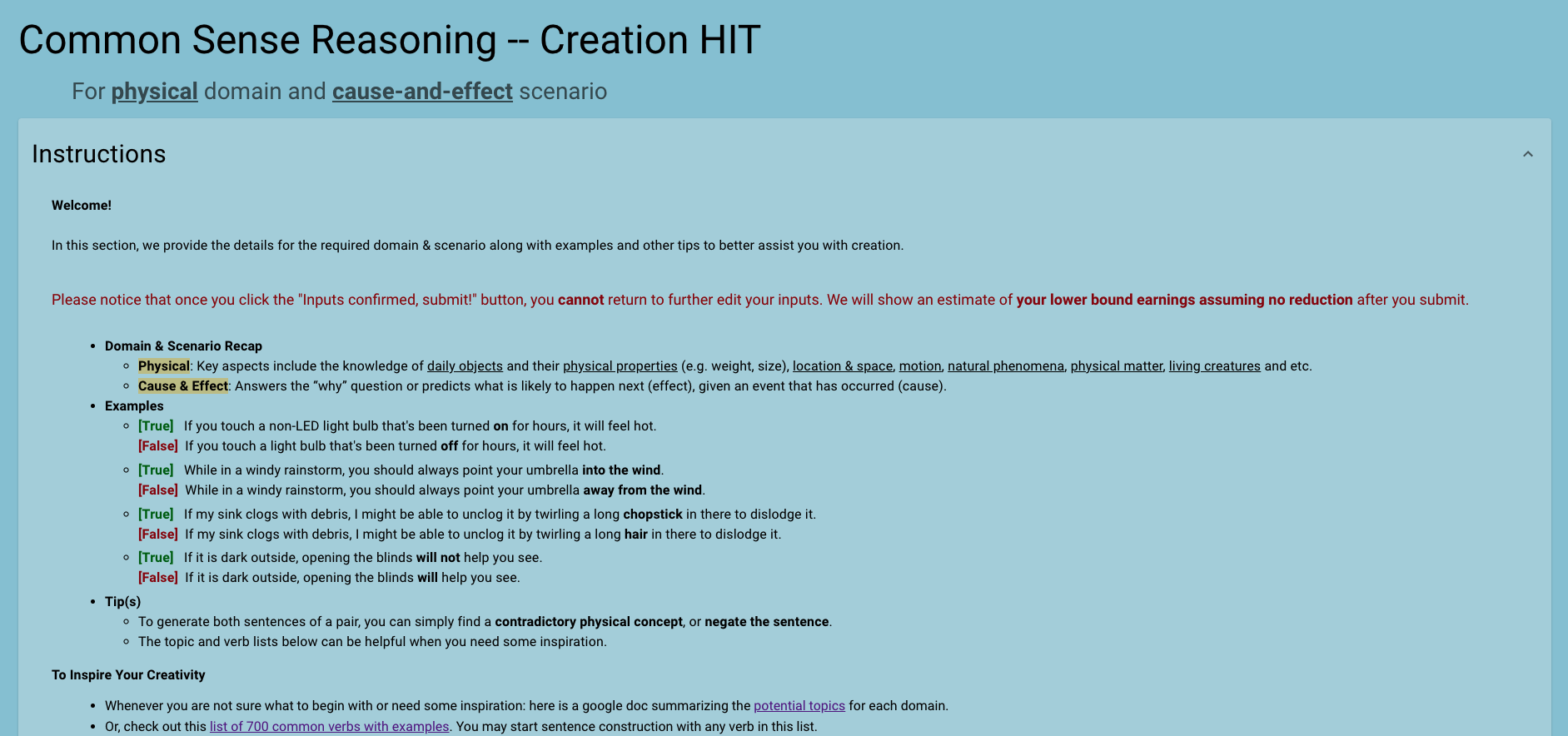}
    \caption{
        Screenshot of the creation interface (instruction section).
    }
    \label{fig:creation_page_1}
    
    \vspace{0.5em}
    
    \centering
    \includegraphics[width=140mm, scale=1]{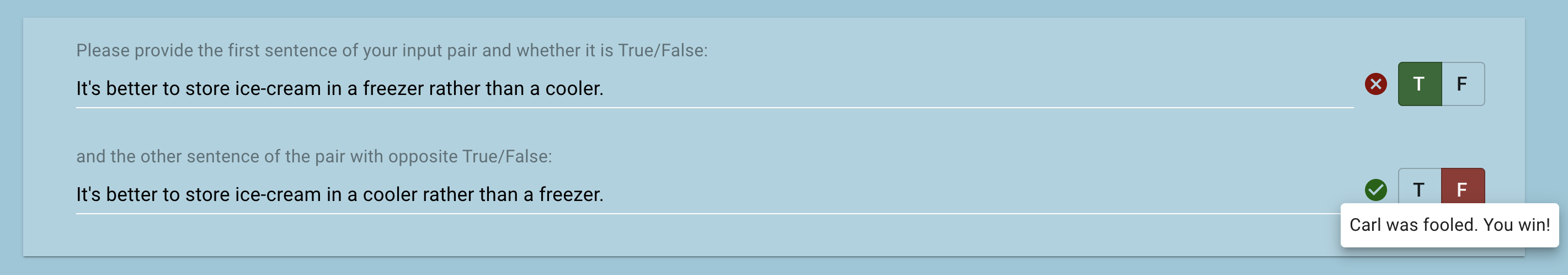}
    \caption{
        Screenshot of the creation interface (1/3 input section).
    }

    \label{fig:creation_page_2}
\end{figure*}

\paragraph{Tool Interface}
Screenshots of our creation interface are as shown in~\figref{fig:creation_page_1} and~\figref{fig:creation_page_2}. We name our deployed model (RoBERTa-large) \textit{Carl} to help emphasize the interactive and gamified creation set-up.

\vspace{.3em}
\mypar{Guidelines}
To inspire workers and collect from more diverse topics of commonsense, we further provide: 1) some hints for having higher chances fooling the model, such as exploiting contradictory physical concepts, negations, swapping entities, etc.,
2) topics pertinent to the domains,
and (3) examples of low quality along with their reasons.

\subsection{Details of Validation}
To ensure data quality, our internal members have helped checking each pair with the validation tool we implement.
For each pair received from the workers, both labels for the statements and their intended domains and scenarios are carefully verified.
For statements which are ambiguous even for humans, if they can be easily fixed by adding more context or better word choices, another round of editing is conducted.

\subsection{Adversarial Setting}
\label{asec:adv}



\begin{figure*}[t!]
    \begin{subtable}{\columnwidth}
    \centering
      \includegraphics[width=\columnwidth]{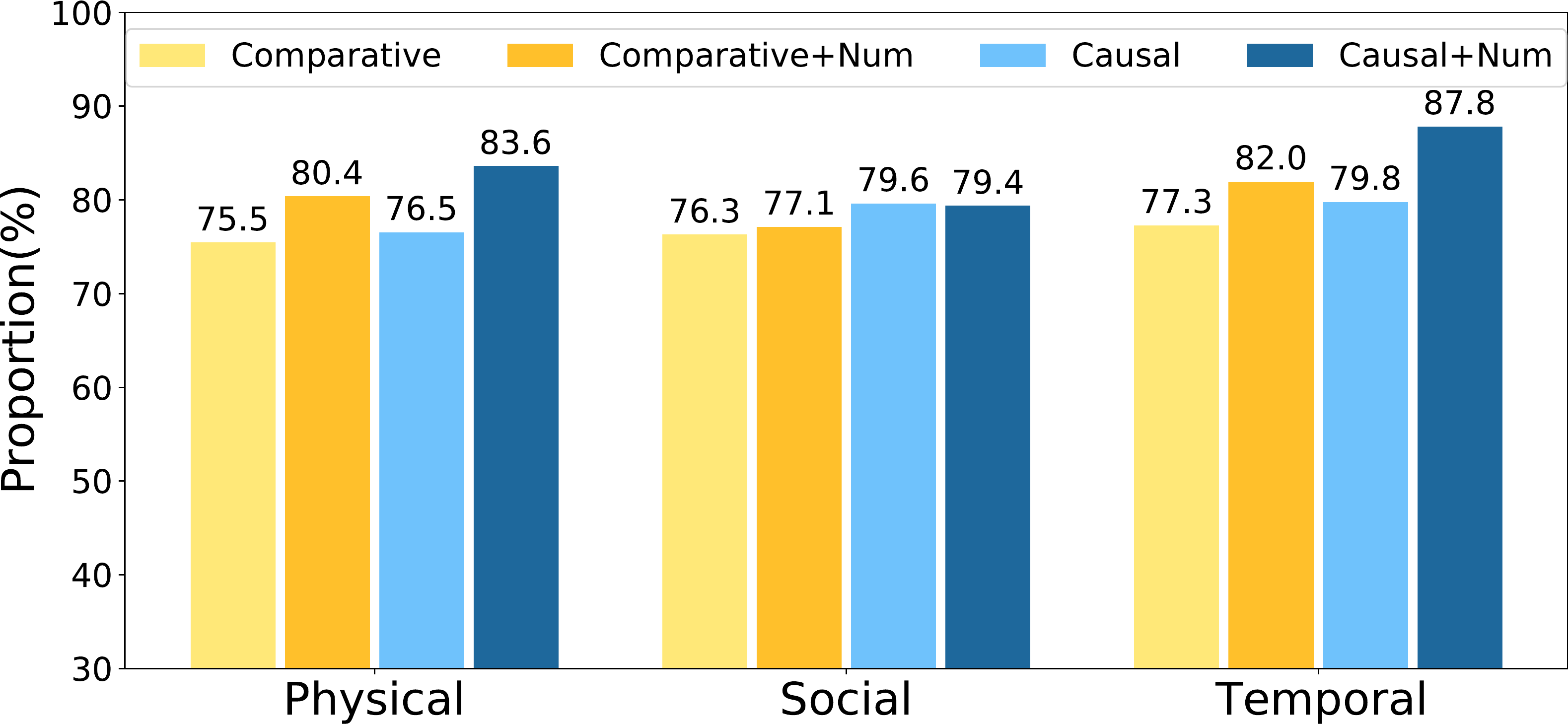}
        \caption{
            Standard
        }
        \label{fig:fool_rate_std}
    \end{subtable}%
\quad
    \begin{subtable}{\columnwidth}
    \centering
      \includegraphics[width=\columnwidth]{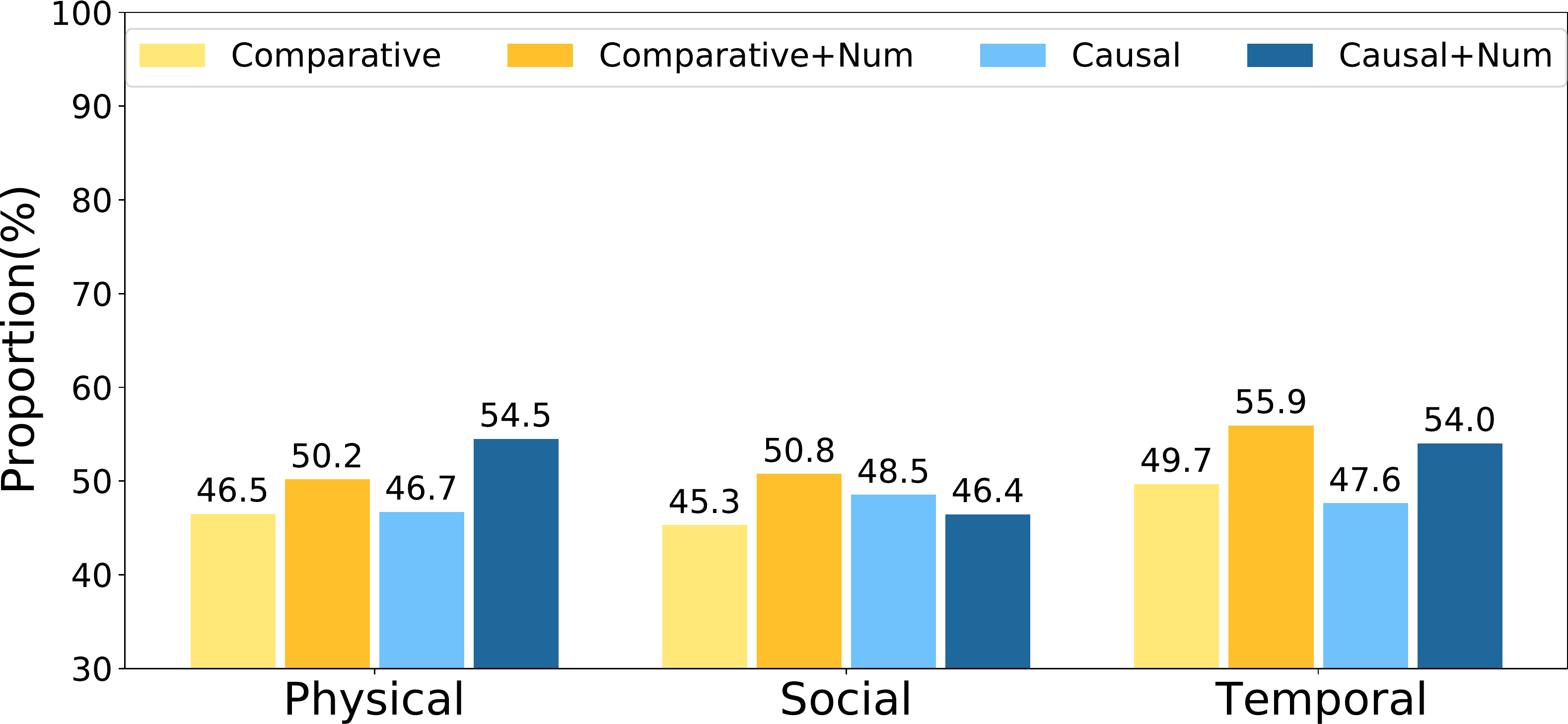}
        \caption{
            Pairwise
        }
        \label{fig:fool_rate_pairwise}
    \end{subtable}
    
\caption{Sentence fooling rates for all categories, \texttt{"+Num"} denotes numeracy involved.}
\label{fig:fooling_rate_all_categories}
\end{figure*}

The total number of collected complementary pairs is around 4.8k, where around .8k are discarded for not having sufficiently high quality, 
\eg "\textit{Frank traded a stock an hour late and lost 80 million dollars.}" and "\textit{Frank traded a stock a second late and lost 80 million dollars.}" Among all the data we collected, the overall fooling rate is 48.55\% per sentence and 78.7\% per pair. For category-specific fooling rates, please refer to ~\figref{fig:fooling_rate_all_categories}.

For sentences that successfully fool the model, we report the mean time of fooling one sentence to be 3.40, the standard deviation (std) as 2.48, and the median as 2.57 (all in minutes). Please notice that the total time is directly retrieved from MTurk and is likely to be overestimated due to worker inactivity. The mean of the number of revisions per fooling sentence is 1.36 with a std as 3.95, and a median as 0 (fooling without re-attempts, requiring no revision). Noticeably, 63\% of the fooling sentences are submitted with no revision.

\begin{figure*}[t!]
    \begin{subtable}{\columnwidth}
    \centering
      \includegraphics[width=\columnwidth]{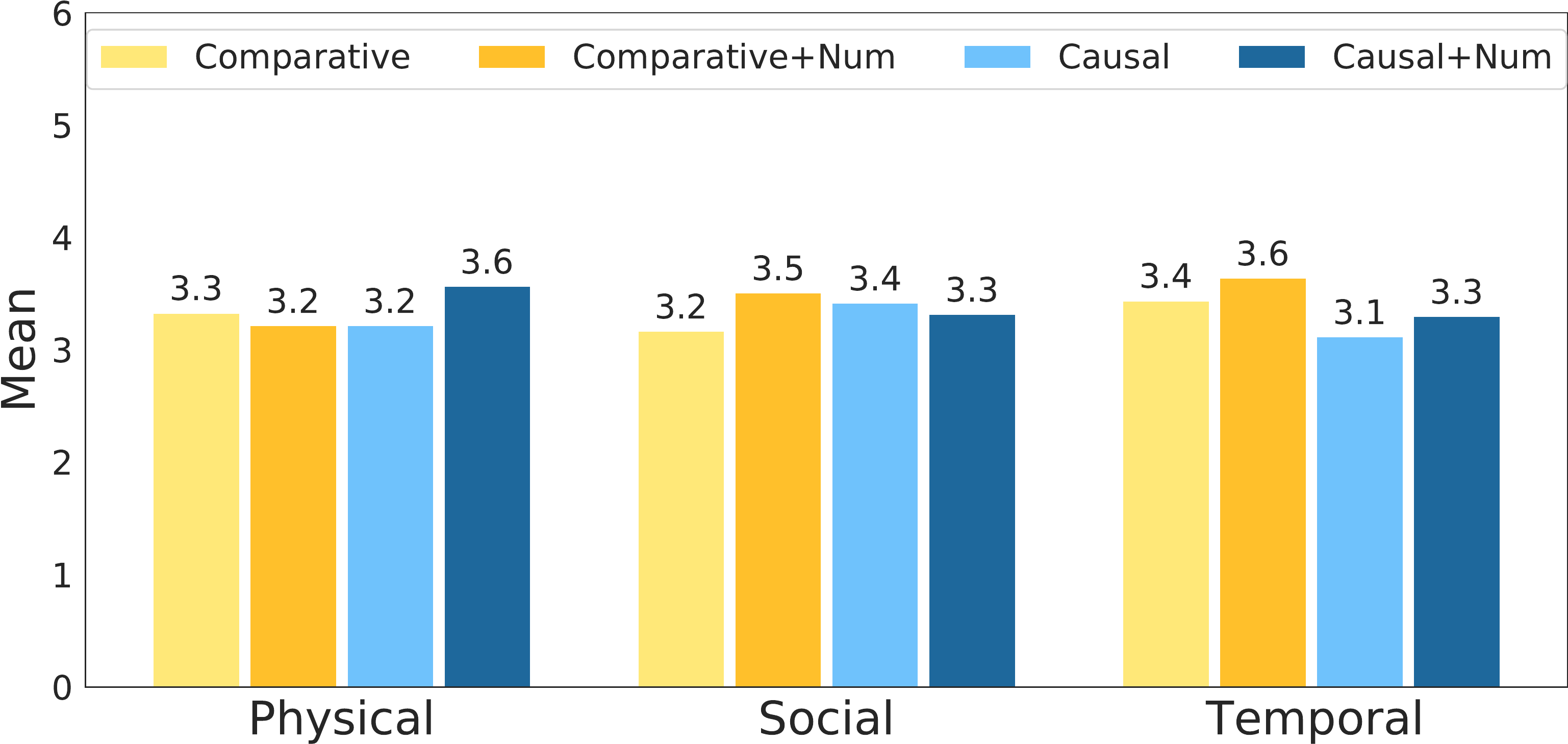}
        \caption{
            Mean
        }
        \label{fig:time_mean}
    \end{subtable}%
\quad
    \begin{subtable}{\columnwidth}
    \centering
      \includegraphics[width=\columnwidth]{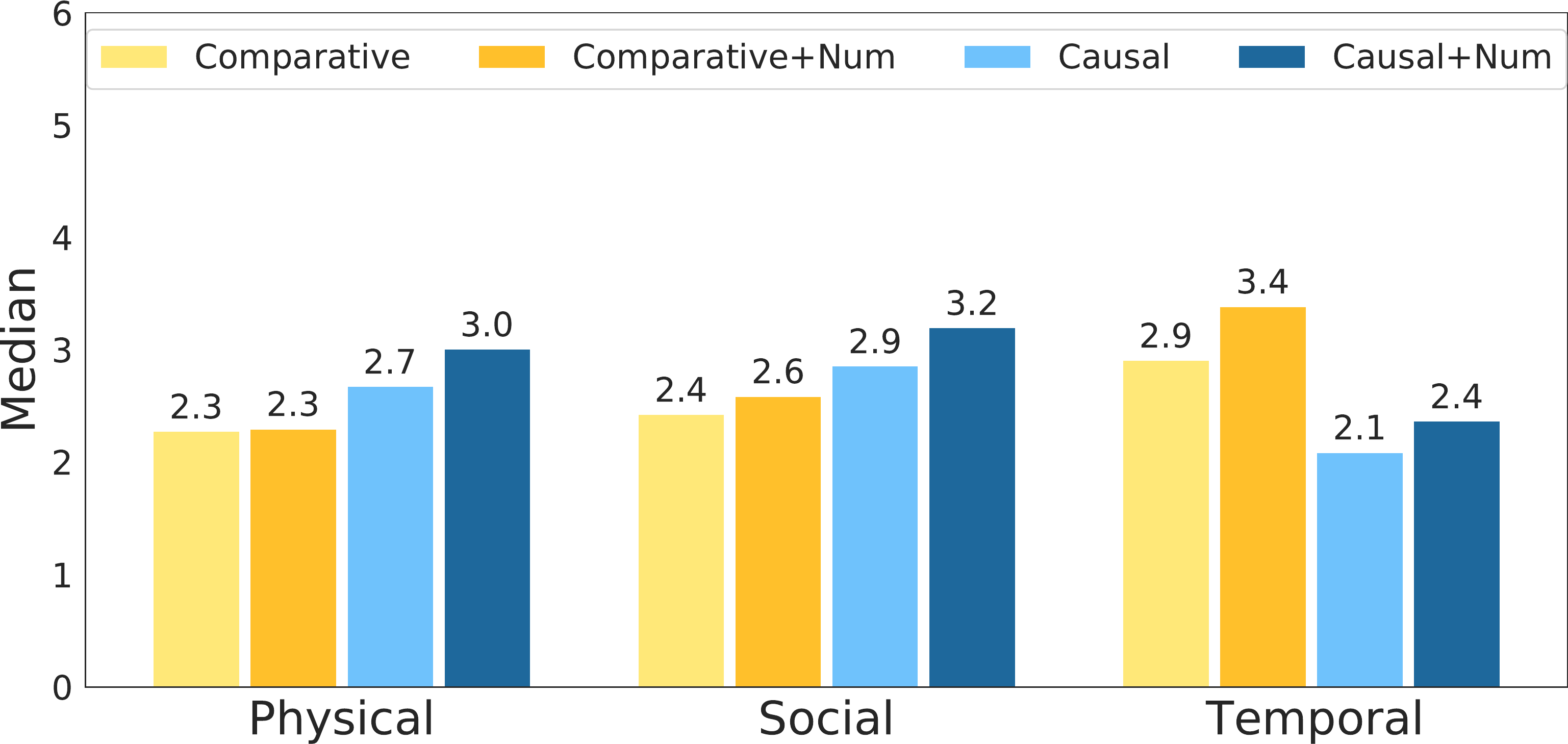}
        \caption{
            Median
        }
        \label{fig:time_median}
    \end{subtable}
    
\caption{Mean and median of the time needed (in minutes) to fool a sentence for all categories, \texttt{"+Num"} denotes numeracy involved.}
\label{fig:time_all_categories}
\end{figure*}

\begin{figure*}[t!]
    \begin{subtable}{\columnwidth}
    \centering
      \includegraphics[width=\columnwidth]{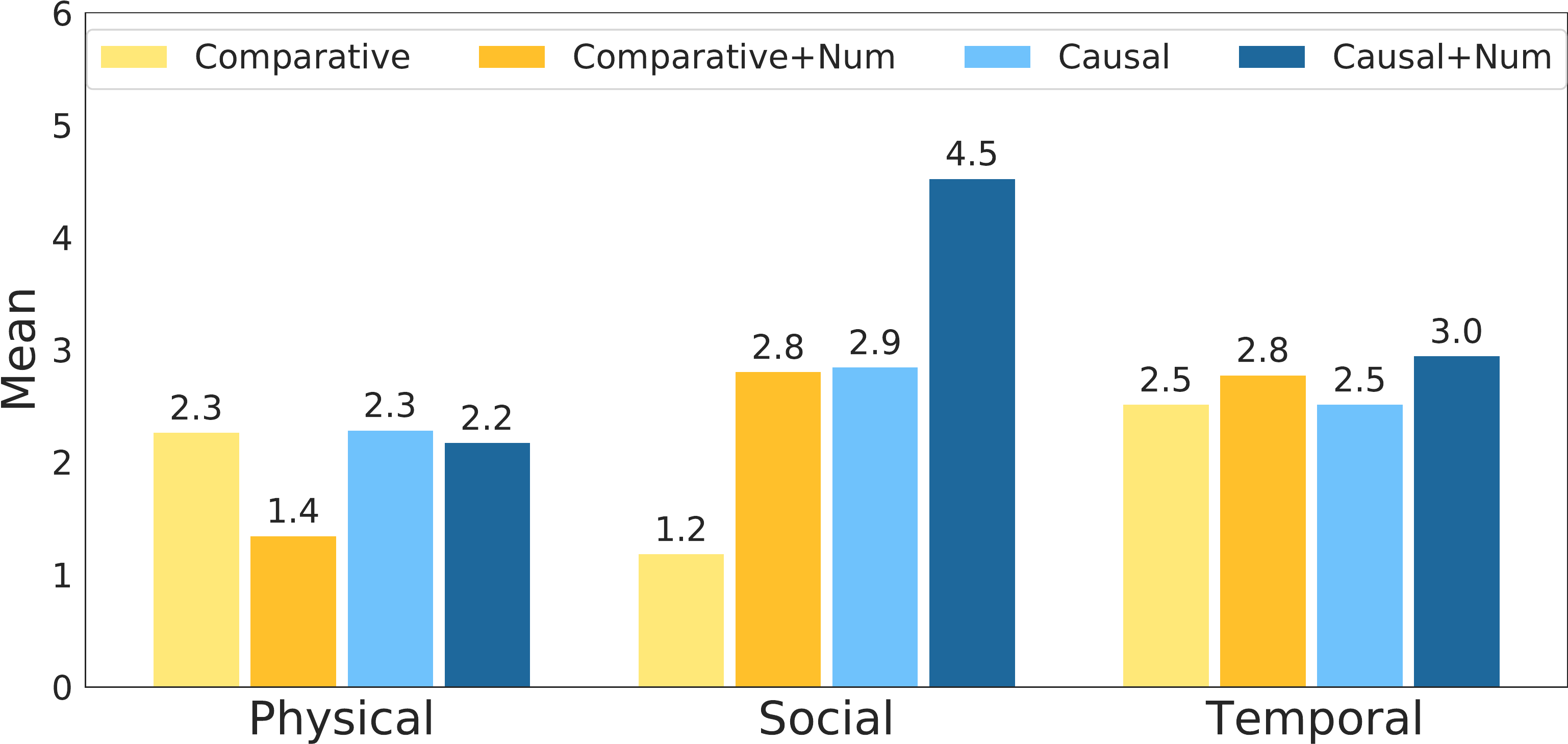}
        \caption{
            Mean
        }
        \label{fig:revision_mean}
    \end{subtable}%
\quad
    \begin{subtable}{\columnwidth}
    \centering
      \includegraphics[width=\columnwidth]{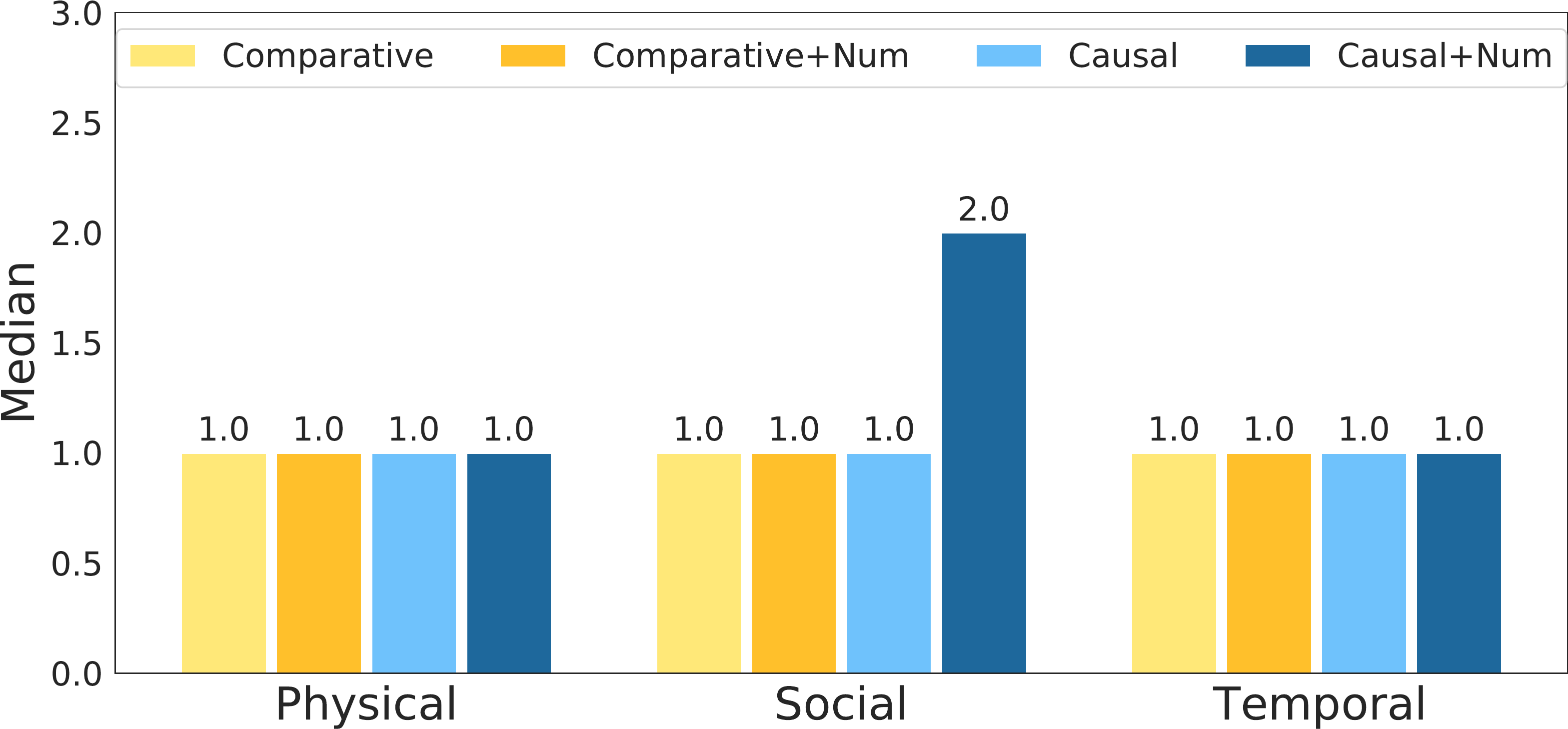}
        \caption{
            Median
        }
        \label{fig:revision_median}
    \end{subtable}
    
\caption{Number of attempts (\ie \# revisions + 1) per sentence for all categories, \texttt{"+Num"} denotes numeracy involved.}
\label{fig:revision_all_categories}
\end{figure*}

For any potential interests,~\figref{fig:time_all_categories} shows the mean and median of required time of fooling per sentence across categories, and similarly ~\figref{fig:revision_all_categories} for the mean and median of the number of attempts which equals to the number of revisions $+1$.

\begin{figure*}
    \centering
    \includegraphics[width=160mm, scale=1]{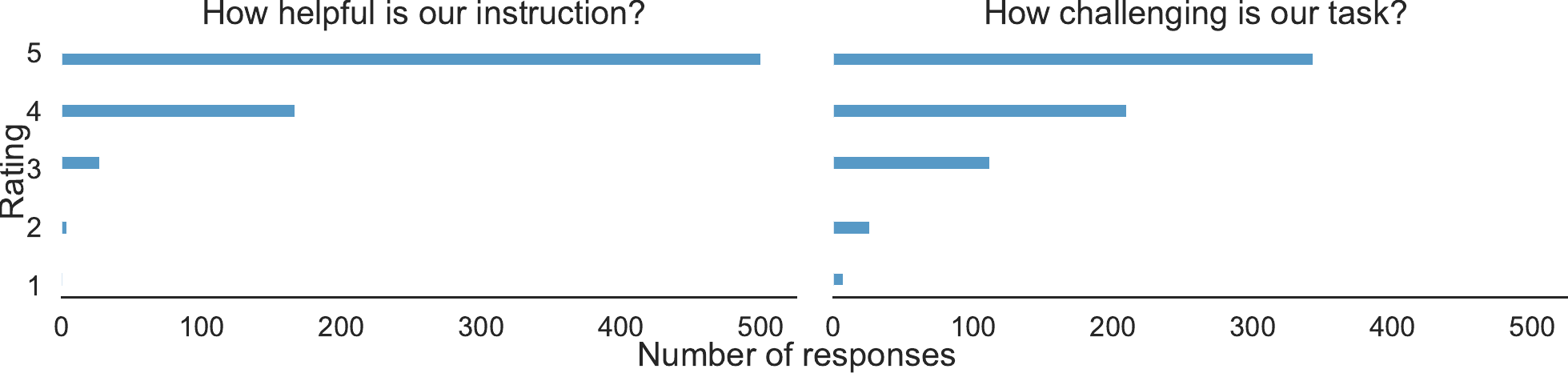}
    \caption{
        Rating distribution in exit survey.
    }
    \label{fig:exit_survey_hist}
\end{figure*}

\figref{fig:exit_survey_hist} shows the distribution of ratings during the exit survey from a total of 699 valid responses.
The survey questions include: 1) how helpful is our instruction?
and 2): how challenging is our task?
For question 1, the mean rating is $4.66\pm0.59$ and median is 5; for question 2, the mean rating is $4.23\pm0.92$ and the median is 4, where 1-5 is from low-to-high rating.

    


\subsection{Statistics of Workers}
\begin{figure}
    \includegraphics[width=70mm, scale=0.5]{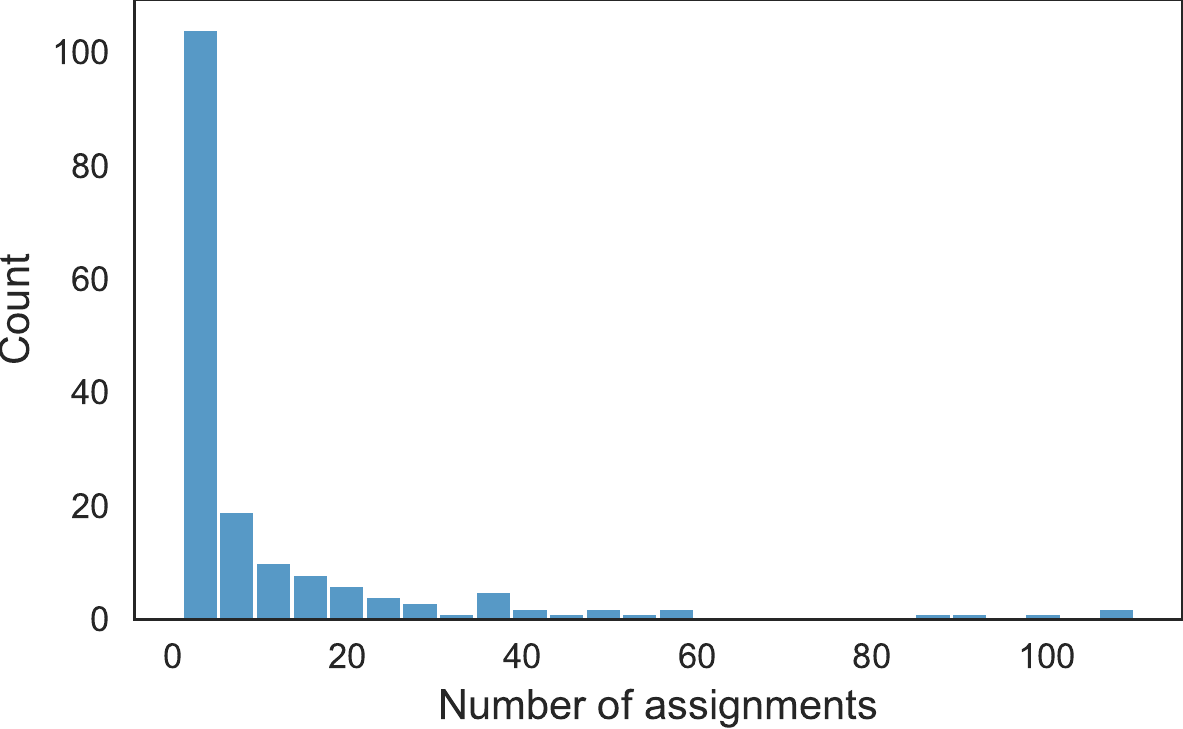}
    \vspace{-1em} 
    \caption{
         Worker counts over the different numbers of assignments.
    }
    \label{fig:num_assignment_per_worker}
    
\end{figure}

\begin{figure}
    \includegraphics[width=75mm, scale=0.7]{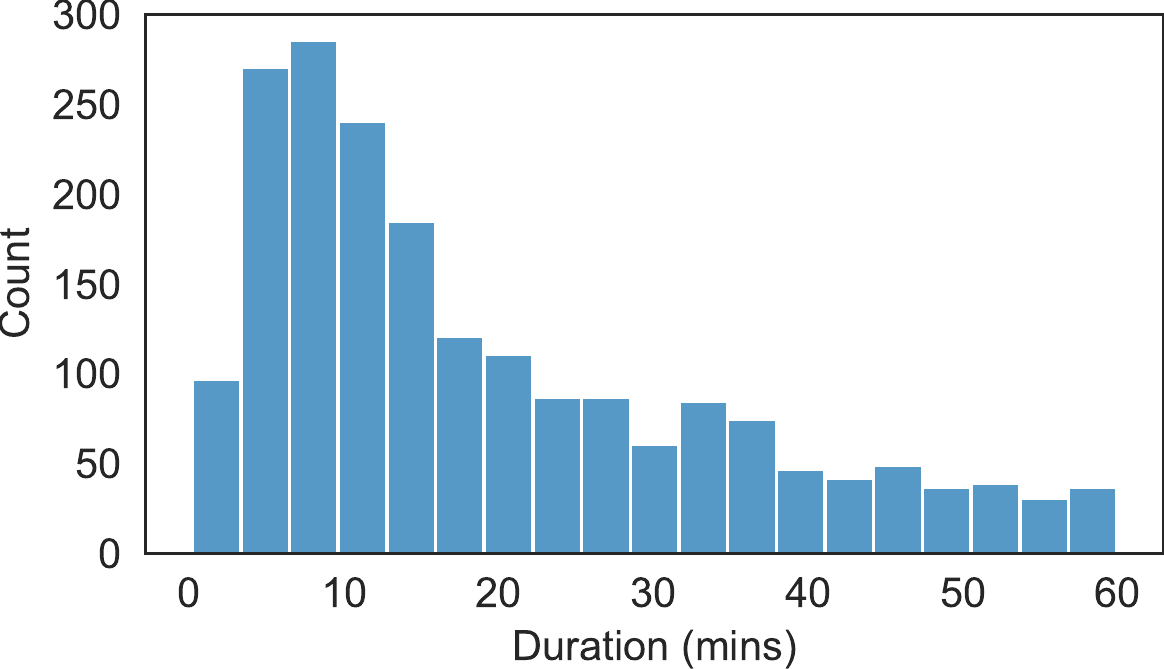}
    \vspace{-1em} 
    \caption{
         Worker counts over different assignment duration (in minutes).
    }
    \label{fig:time_per_assignment}
\end{figure}

173 workers participated in our task, and~\figref{fig:num_assignment_per_worker} shows the worker counts for the different numbers of assignments attempted by each of the workers, and~\figref{fig:time_per_assignment} shows the worker counts for the time (duration, in minutes) each worker spent on one assignment. 


\section{Additional Details on Baseline Models}
\label{asec:baselines}

We include several essential implementation details of the benchmark models in the following:

\vspace{.3em}
\mypar{Bi-LSTM+GloVe} Our Bi-LSTM model~\cite{hochreiter1997long} is one-layered with a 512-dimensional hidden layer, which takes input word embeddings from 300-dimensional GloVe word embeddings~\cite{pennington-etal-2014-glove}. We train all LSTM layers from scratch.

\vspace{.3em}

\mypar{BERT-base Models}
For BERT-style architectures we employ a multi-layer-perceptron (MLP) on top of the [CLS] special token for binary prediction. 
\vspace{.3em}

\mypar{T5-large}
To adopt T5-large's text-to-text format to our dataset, we use the prefix \textit{com2sense sentence:} and the labels \textit{True} and \textit{False} as model output.
\vspace{.3em}

\mypar{UnifiedQA Models}
We use two UnifiedQA Models. One with the T5-large backbone and one with the T5-3b backbone. For these models, we use \textit{Is the following sentence correct?} as the prefix, to create a question. Then as the answer we use \textit{Yes} / \textit{No}.
\section{More Details on the Experiments}
\label{asec:training}

\subsection{Hyperparameters}

\begin{table*}[t]
\centering
    \scalebox{.7}{
\begin{tabular}{cccccccc}
\hline
\multirow{2}{*}{\textbf{Model}} & \multirow{2}{*}{\textbf{\# Params}} & \multirow{2}{*}{\textbf{Batch-Size}} & \multirow{2}{*}{\textbf{LR}} & \multirow{2}{*}{\textbf{\begin{tabular}[c]{@{}c@{}}Training\\  Iterations\end{tabular}}} & \multirow{2}{*}{\textbf{\begin{tabular}[c]{@{}c@{}}Gradient Accumulation\\ Steps\end{tabular}}} & \multirow{2}{*}{\textbf{\begin{tabular}[c]{@{}c@{}}Max. Token \\ Length\end{tabular}}} & \multirow{2}{*}{\textbf{}} \\
 &  &  &  &  &  &  &  \\ \hline
BiLSTM+GloVe & 3.5M & 64 & $1 \times 10^{-5}$ & 100 & 4 & 80 &  \\ \hline
BERT-base & 109.5M & 64 & $1 \times 10^{-5}$ & 100 & 4 & 80 &  \\ \hline
RoBERTa-large & 355.4M & 32 & $1 \times 10^{-5}$ & 100 & 4 & 80 &  \\ \hline
DeBERTa-large & 405.2M & 32 & $1 \times 10^{-5}$ & 100 & 4 & 80 &  \\ \hline
T5-large & 737.5M & 8 & $1 \times 10^{-5}$ & 100 & 4 & 80 &  \\ \hline
UnifiedQA-t5-large & 737.5M & 8 & $1 \times 10^{-5}$ & 100 & 4 & 80 &  \\ \hline
UnifiedQA-t5-3b & 3000M & 2 & $1 \times 10^{-5}$ & 100 & 8 & 64 &  \\ \hline
 &  &  &  &  &  &  &  \\ \cline{2-6}
 & Bound (lower-upper) & 1-64 & $5 \times 10^{-5}$--$1 \times 10^{-6}$ & 10-100 & 1-10 &  &  \\ \cline{2-6}
 & Number of trials & 2-3 & 2-3 & 2-3 & 2-3 &  &  \\ \cline{2-6}
\end{tabular}
}
\caption{Hyperparameters used for each model during finetuning on~\papername{} along with the search bounds for them: \textit{LR} denotes the learning rate that does not change during the training process.  All the models are trained with Adam optimizers~\cite{kingma2014adam}. We include number of parameters of each model in the first column, denoted as \textit{\# params}.}
\label{tab:hyparams}
\end{table*}




All the essential hyperparameters used throughout this work can be referred to in~\tbref{tab:hyparams}.
We also include the search bounds as well as the number of trials in searching for our manually-tuned hyperparameter search procedures in~\tbref{tab:hyparams}.

\subsection{Validation Set Results}

We validate all trained models on a 402-pair validation set and tune the hyperparameters accordingly. The performances on the validation set are reported in ~\tbref{tab:baseline_val_results}.
\begingroup
\setlength{\tabcolsep}{12pt} 

\begin{table}[h!]
	\centering
		\small
	\begin{tabular}{lcc}
		\toprule
		\textbf{Model}  & Standard  & Pairwise \\
		\midrule
		Random          & 50.00     & 25.00 \\
		BiLSTM+GloVe    & 52.80    & 27.50 \\
		BERT-base       & 57.07     & 23.11 \\   
		RoBERTa-large   & 62.81     & 38.30 \\
		T5-large        & 62.81     & 35.82 \\   
		UnifiedQA-large & 63.43     & 37.31 \\
		DeBERTa-large   & 66.29     & 43.03 \\  
		UnifiedQA-3b & \textbf{75.12} & \textbf{56.22} \\
    
		\bottomrule
	\end{tabular}
	\caption{Validation-set accuracy for selected models, trained and evaluated on respective datasets.
}
	\label{tab:baseline_val_results}
\end{table}
\endgroup

\vspace{-0.5em}
\subsection{Performance Across Input Lengths}
\begin{figure}
    \centering
    \includegraphics[width=60mm, height=45mm,
    scale=0.5]{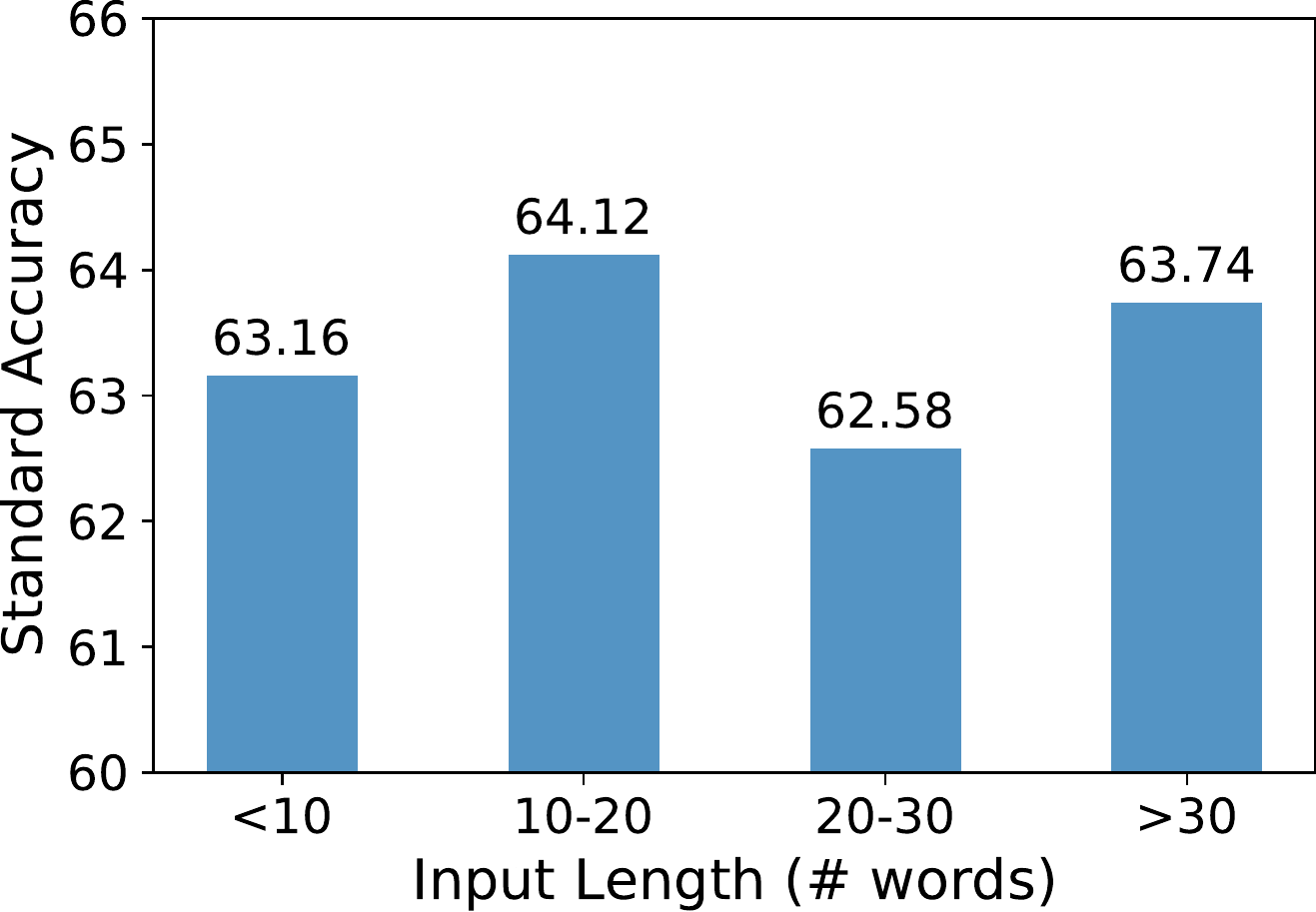}
    \caption{
        Standard accuracy of the DeBERTa-large model measured on subsets of data with different input lengths.
    }
    \vspace{-1em}
    \label{fig:acc-vs-length}
\end{figure}

Although the sentence length in our dataset varies, we find no obvious relation between the length of the sentences and the difficulty for the model to comprehend, in terms of accuracy. As depicted in~\figref{fig:acc-vs-length}, we therefore conclude that sentence length would not have significant influence on fooling models including DeBERTa-large.

\subsection{Software, Hardware, \& Other Details}

Transformer-based models are implemented via the HuggingFace PyTorch API ~\cite{wolf-etal-2020-transformers}.
All the benchmarked models, except for UnifiedQA-T5-3b are trained on Nvidia GeForce 2080Ti GPUs\footnote{https://www.nvidia.com/en-us/geforce/graphics-cards/rtx-2080-ti/} on a CentOS 7 operating system. The UnifiedQA-T5-3b is trained on NVIDIA Tesla V100 GPUs \footnote{https://www.nvidia.com/en-gb/data-center/tesla-v100/} on an Ubuntu 18 operating system.
The T5-large and UnifiedQA-T5-large are trained using the model parallelism approach on two GPUs. The UnifiedQA-T5-3b is trained using model parallelism on 8 GPUS.

The maximum training time is approximately 6 hours for all the models, with the BERT-style models on the lower end of the range and the T5-style models on the higher end.

%% file: main.bbl
\begin{thebibliography}{50}
\expandafter\ifx\csname natexlab\endcsname\relax\def\natexlab#1{#1}\fi

\bibitem[{Bartolo et~al.(2020)Bartolo, Roberts, Welbl, Riedel, and
  Stenetorp}]{bartolo-etal-2020-beat}
Max Bartolo, Alastair Roberts, Johannes Welbl, Sebastian Riedel, and Pontus
  Stenetorp. 2020.
\newblock \href {https://doi.org/10.1162/tacl_a_00338} {Beat the {AI}:
  Investigating adversarial human annotation for reading comprehension}.
\newblock In \emph{Transactions of the Association for Computational
  Linguistics (TACL)}, volume~8, pages 662--678.

\bibitem[{Bhagavatula et~al.(2020)Bhagavatula, Bras, Malaviya, Sakaguchi,
  Holtzman, Rashkin, Downey, tau Yih, and Choi}]{bhagavatula2020abductive}
Chandra Bhagavatula, Ronan~Le Bras, Chaitanya Malaviya, Keisuke Sakaguchi, Ari
  Holtzman, Hannah Rashkin, Doug Downey, Wen tau Yih, and Yejin Choi. 2020.
\newblock \href {https://openreview.net/forum?id=Byg1v1HKDB} {Abductive
  commonsense reasoning}.
\newblock In \emph{International Conference on Learning Representations
  (ICLR)}.

\bibitem[{Bisk et~al.(2020)Bisk, Zellers, Bras, Gao, and Choi}]{Bisk2020PIQARA}
Yonatan Bisk, Rowan Zellers, Ronan~Le Bras, Jianfeng Gao, and Yejin Choi. 2020.
\newblock Piqa: Reasoning about physical commonsense in natural language.
\newblock In \emph{Proceedings of the National Conference on Artificial
  Intelligence (AAAI)}.

\bibitem[{Chen et~al.(2019)Chen, D{'}Arcy, Liu, Fernandez, and
  Downey}]{chen-etal-2019-codah}
Michael Chen, Mike D{'}Arcy, Alisa Liu, Jared Fernandez, and Doug Downey. 2019.
\newblock \href {https://doi.org/10.18653/v1/W19-2008} {{CODAH}: An
  adversarially-authored question answering dataset for common sense}.
\newblock In \emph{Proceedings of the 3rd Workshop on Evaluating Vector Space
  Representations for {NLP}}, pages 63--69, Minneapolis, USA. Association for
  Computational Linguistics.

\bibitem[{Clark et~al.(2019)Clark, Lee, Chang, Kwiatkowski, Collins, and
  Toutanova}]{clark2019boolq}
Christopher Clark, Kenton Lee, Ming-Wei Chang, Tom Kwiatkowski, Michael
  Collins, and Kristina Toutanova. 2019.
\newblock {B}ool{Q}: Exploring the surprising difficulty of natural yes/no
  questions.
\newblock In \emph{Proceedings of the North American Chapter of the Association
  for Computational Linguistics - Human Language Technologies (NAACL-HLT)},
  pages 2924--2936, Minneapolis, Minnesota. Association for Computational
  Linguistics.

\bibitem[{Devlin et~al.(2019)Devlin, Chang, Lee, and
  Toutanova}]{devlin-etal-2019-bert}
Jacob Devlin, Ming-Wei Chang, Kenton Lee, and Kristina Toutanova. 2019.
\newblock \href {https://doi.org/10.18653/v1/N19-1423} {{BERT}: Pre-training of
  deep bidirectional transformers for language understanding}.
\newblock In \emph{Proceedings of the North American Chapter of the Association
  for Computational Linguistics - Human Language Technologies (NAACL-HLT)},
  pages 4171--4186, Minneapolis, Minnesota. Association for Computational
  Linguistics.

\bibitem[{Dua et~al.(2019)Dua, Wang, Dasigi, Stanovsky, Singh, and
  Gardner}]{dua-etal-2019-drop}
Dheeru Dua, Yizhong Wang, Pradeep Dasigi, Gabriel Stanovsky, Sameer Singh, and
  Matt Gardner. 2019.
\newblock \href {https://doi.org/10.18653/v1/N19-1246} {{DROP}: A reading
  comprehension benchmark requiring discrete reasoning over paragraphs}.
\newblock In \emph{Proceedings of the North American Chapter of the Association
  for Computational Linguistics - Human Language Technologies (NAACL-HLT)},
  pages 2368--2378, Minneapolis, Minnesota. Association for Computational
  Linguistics.

\bibitem[{Gardner et~al.(2020)Gardner, Artzi, Basmov, Berant, Bogin, Chen,
  Dasigi, Dua, Elazar, Gottumukkala et~al.}]{gardner2020evaluating}
Matt Gardner, Yoav Artzi, Victoria Basmov, Jonathan Berant, Ben Bogin, Sihao
  Chen, Pradeep Dasigi, Dheeru Dua, Yanai Elazar, Ananth Gottumukkala, et~al.
  2020.
\newblock Evaluating models’ local decision boundaries via contrast sets.
\newblock In \emph{Proceedings of the 2020 Conference on Empirical Methods in
  Natural Language Processing: Findings}, pages 1307--1323.

\bibitem[{Geva et~al.(2019)Geva, Goldberg, and Berant}]{GevaEtAl2019}
Mor Geva, Yoav Goldberg, and Jonathan Berant. 2019.
\newblock \href {https://doi.org/10.18653/v1/D19-1107} {Are we modeling the
  task or the annotator? an investigation of annotator bias in natural language
  understanding datasets}.
\newblock In \emph{Proceedings of the Conference on Empirical Methods for
  Natural Language Processing (EMNLP)}, pages 1161--1166, Hong Kong, China.
  Association for Computational Linguistics.

\bibitem[{Glockner et~al.(2018)Glockner, Shwartz, and
  Goldberg}]{glockner2018breaking}
Max Glockner, Vered Shwartz, and Yoav Goldberg. 2018.
\newblock Breaking nli systems with sentences that require simple lexical
  inferences.
\newblock In \emph{Proceedings of the Annual Meeting of the Association of
  Computational Linguistics (ACL)}.

\bibitem[{Gururangan et~al.(2018)Gururangan, Swayamdipta, Levy, Schwartz,
  Bowman, and Smith}]{gururangan-etal-2018-annotation}
Suchin Gururangan, Swabha Swayamdipta, Omer Levy, Roy Schwartz, Samuel Bowman,
  and Noah~A. Smith. 2018.
\newblock \href {https://doi.org/10.18653/v1/N18-2017} {Annotation artifacts in
  natural language inference data}.
\newblock In \emph{Proceedings of the North American Chapter of the Association
  for Computational Linguistics - Human Language Technologies (NAACL-HLT)},
  pages 107--112, New Orleans, Louisiana. Association for Computational
  Linguistics.

\bibitem[{He et~al.(2020)He, Liu, Gao, and Chen}]{he2020deberta}
Pengcheng He, Xiaodong Liu, Jianfeng Gao, and Weizhu Chen. 2020.
\newblock Deberta: Decoding-enhanced bert with disentangled attention.
\newblock \emph{arXiv preprint arXiv:2006.03654}.

\bibitem[{Hochreiter and Schmidhuber(1997)}]{hochreiter1997long}
Sepp Hochreiter and J{\"u}rgen Schmidhuber. 1997.
\newblock Long short-term memory.
\newblock In \emph{Neural computation}, volume~9, pages 1735--1780. MIT Press.

\bibitem[{Huang et~al.(2019)Huang, Bras, Bhagavatula, and
  Choi}]{huang2019cosmos}
Lifu Huang, Ronan~Le Bras, Chandra Bhagavatula, and Yejin Choi. 2019.
\newblock Cosmos qa: Machine reading comprehension with contextual commonsense
  reasoning.
\newblock In \emph{Proceedings of the Conference on Empirical Methods for
  Natural Language Processing (EMNLP)}.

\bibitem[{Jia and Liang(2017)}]{jia2017adversarial}
Robin Jia and Percy Liang. 2017.
\newblock Adversarial examples for evaluating reading comprehension systems.
\newblock In \emph{Proceedings of the Conference on Empirical Methods for
  Natural Language Processing (EMNLP)}.

\bibitem[{Kaushik et~al.(2020)Kaushik, Hovy, and Lipton}]{kaushik2019learning}
Divyansh Kaushik, Eduard Hovy, and Zachary~C Lipton. 2020.
\newblock Learning the difference that makes a difference with
  counterfactually-augmented data.
\newblock In \emph{International Conference on Learning Representations
  (ICLR)}.

\bibitem[{Kaushik and Lipton(2018)}]{kaushik-lipton-2018-much}
Divyansh Kaushik and Zachary~C. Lipton. 2018.
\newblock \href {https://doi.org/10.18653/v1/D18-1546} {How much reading does
  reading comprehension require? a critical investigation of popular
  benchmarks}.
\newblock In \emph{Proceedings of the Conference on Empirical Methods for
  Natural Language Processing (EMNLP)}, pages 5010--5015, Brussels, Belgium.
  Association for Computational Linguistics.

\bibitem[{Khashabi et~al.(2020)Khashabi, Min, Khot, Sabharwal, Tafjord, Clark,
  and Hajishirzi}]{khashabi2020unifiedqa}
Daniel Khashabi, Sewon Min, Tushar Khot, Ashish Sabharwal, Oyvind Tafjord,
  Peter Clark, and Hannaneh Hajishirzi. 2020.
\newblock {U}nified{QA}: Crossing format boundaries with a single qa system.
\newblock In \emph{Findings of the Association for Computational Linguistics:
  EMNLP}.

\bibitem[{Kingma and Ba(2015)}]{kingma2014adam}
Diederik~P Kingma and Jimmy Ba. 2015.
\newblock Adam: A method for stochastic optimization.
\newblock In \emph{International Conference on Learning Representations
  (ICLR)}.

\bibitem[{Le~Bras et~al.(2020)Le~Bras, Swayamdipta, Bhagavatula, Zellers,
  Peters, Sabharwal, and Choi}]{le2020adversarial}
Ronan Le~Bras, Swabha Swayamdipta, Chandra Bhagavatula, Rowan Zellers, Matthew
  Peters, Ashish Sabharwal, and Yejin Choi. 2020.
\newblock Adversarial filters of dataset biases.
\newblock In \emph{Proceedings of the International Conference on Machine
  Learning (ICML)}, pages 1078--1088. PMLR.

\bibitem[{Levesque et~al.(2012)Levesque, Davis, and
  Morgenstern}]{Levesque2011TheWS}
H.~Levesque, E.~Davis, and L.~Morgenstern. 2012.
\newblock The winograd schema challenge.
\newblock In \emph{Proceedings of the International Conference on Principles of
  Knowledge Representation and Reasoning (KR)}.

\bibitem[{Lin et~al.(2020{\natexlab{a}})Lin, Lee, Khanna, and
  Ren}]{lin-etal-2020-birds}
Bill~Yuchen Lin, Seyeon Lee, Rahul Khanna, and Xiang Ren. 2020{\natexlab{a}}.
\newblock \href {https://doi.org/10.18653/v1/2020.emnlp-main.557} {{B}irds have
  four legs?! {N}umer{S}ense: {P}robing {N}umerical {C}ommonsense {K}nowledge
  of {P}re-{T}rained {L}anguage {M}odels}.
\newblock In \emph{Proceedings of the Conference on Empirical Methods for
  Natural Language Processing (EMNLP)}, pages 6862--6868, Online. Association
  for Computational Linguistics.

\bibitem[{Lin et~al.(2020{\natexlab{b}})Lin, Zhou, Shen, Zhou, Bhagavatula,
  Choi, and Ren}]{lin-etal-2020-commongen}
Bill~Yuchen Lin, Wangchunshu Zhou, Ming Shen, Pei Zhou, Chandra Bhagavatula,
  Yejin Choi, and Xiang Ren. 2020{\natexlab{b}}.
\newblock \href {https://doi.org/10.18653/v1/2020.findings-emnlp.165}
  {{C}ommon{G}en: A constrained text generation challenge for generative
  commonsense reasoning}.
\newblock In \emph{Findings of the Association for Computational Linguistics:
  EMNLP}, pages 1823--1840, Online. Association for Computational Linguistics.

\bibitem[{Liu et~al.(2019)Liu, Ott, Goyal, Du, Joshi, Chen, Levy, Lewis,
  Zettlemoyer, and Stoyanov}]{liu2019roberta}
Yinhan Liu, Myle Ott, Naman Goyal, Jingfei Du, Mandar Joshi, Danqi Chen, Omer
  Levy, Mike Lewis, Luke Zettlemoyer, and Veselin Stoyanov. 2019.
\newblock Roberta: A robustly optimized bert pretraining approach.
\newblock \emph{arXiv preprint arXiv:1907.11692}.

\bibitem[{McCoy et~al.(2019)McCoy, Pavlick, and Linzen}]{mccoy-etal-2019-right}
Tom McCoy, Ellie Pavlick, and Tal Linzen. 2019.
\newblock \href {https://doi.org/10.18653/v1/P19-1334} {Right for the wrong
  reasons: Diagnosing syntactic heuristics in natural language inference}.
\newblock In \emph{Proceedings of the Annual Meeting of the Association of
  Computational Linguistics (ACL)}, pages 3428--3448, Florence, Italy.
  Association for Computational Linguistics.

\bibitem[{Mostafazadeh et~al.(2016)Mostafazadeh, Chambers, He, Parikh, Batra,
  Vanderwende, Kohli, and Allen}]{mostafazadeh2016corpus}
Nasrin Mostafazadeh, Nathanael Chambers, Xiaodong He, Devi Parikh, Dhruv Batra,
  Lucy Vanderwende, Pushmeet Kohli, and James Allen. 2016.
\newblock A corpus and cloze evaluation for deeper understanding of commonsense
  stories.
\newblock In \emph{Proceedings of the North American Chapter of the Association
  for Computational Linguistics - Human Language Technologies (NAACL-HLT)},
  pages 839--849.

\bibitem[{Nie et~al.(2020)Nie, Williams, Dinan, Bansal, Weston, and
  Kiela}]{nie-etal-2020-adversarial}
Yixin Nie, Adina Williams, Emily Dinan, Mohit Bansal, Jason Weston, and Douwe
  Kiela. 2020.
\newblock \href {https://doi.org/10.18653/v1/2020.acl-main.441} {Adversarial
  {NLI}: A new benchmark for natural language understanding}.
\newblock In \emph{Proceedings of the Annual Meeting of the Association of
  Computational Linguistics (ACL)}, pages 4885--4901, Online. Association for
  Computational Linguistics.

\bibitem[{Ning et~al.(2020)Ning, Wu, Han, Peng, Gardner, and
  Roth}]{ning2020torque}
Qiang Ning, Hao Wu, Rujun Han, Nanyun Peng, Matt Gardner, and Dan Roth. 2020.
\newblock Torque: A reading comprehension dataset of temporal ordering
  questions.
\newblock In \emph{Proceedings of the Conference on Empirical Methods for
  Natural Language Processing (EMNLP)}, pages 1158--1172. Association for
  Computational Linguistics.

\bibitem[{Pennington et~al.(2014)Pennington, Socher, and
  Manning}]{pennington-etal-2014-glove}
Jeffrey Pennington, Richard Socher, and Christopher Manning. 2014.
\newblock \href {https://doi.org/10.3115/v1/D14-1162} {{G}lo{V}e: Global
  vectors for word representation}.
\newblock In \emph{Proceedings of the Conference on Empirical Methods for
  Natural Language Processing (EMNLP)}, pages 1532--1543, Doha, Qatar.
  Association for Computational Linguistics.

\bibitem[{Poliak et~al.(2018)Poliak, Naradowsky, Haldar, Rudinger, and
  Van~Durme}]{poliak-etal-2018-hypothesis}
Adam Poliak, Jason Naradowsky, Aparajita Haldar, Rachel Rudinger, and Benjamin
  Van~Durme. 2018.
\newblock Hypothesis only baselines in natural language inference.
\newblock In \emph{Proceedings of the Seventh Joint Conference on Lexical and
  Computational Semantics}, pages 180--191, New Orleans, Louisiana. Association
  for Computational Linguistics.

\bibitem[{Raffel et~al.(2020)Raffel, Shazeer, Roberts, Lee, Narang, Matena,
  Zhou, Li, and Liu}]{raffel2020exploring}
Colin Raffel, Noam Shazeer, Adam Roberts, Katherine Lee, Sharan Narang, Michael
  Matena, Yanqi Zhou, Wei Li, and Peter~J. Liu. 2020.
\newblock \href {http://jmlr.org/papers/v21/20-074.html} {Exploring the limits
  of transfer learning with a unified text-to-text transformer}.
\newblock In \emph{Journal of Machine Learning Research}, volume~21, pages
  1--67.

\bibitem[{Ravichander et~al.(2019)Ravichander, Naik, Rose, and
  Hovy}]{ravichander-etal-2019-equate}
Abhilasha Ravichander, Aakanksha Naik, Carolyn Rose, and Eduard Hovy. 2019.
\newblock \href {https://doi.org/10.18653/v1/K19-1033} {{EQUATE}: A benchmark
  evaluation framework for quantitative reasoning in natural language
  inference}.
\newblock In \emph{Proceedings of the Annual Conference on Computational
  Natural Language Learning (CoNLL)}, pages 349--361, Hong Kong, China.
  Association for Computational Linguistics.

\bibitem[{Sakaguchi et~al.(2020)Sakaguchi, Bras, Bhagavatula, and
  Choi}]{Sakaguchi2020WINOGRANDEAA}
Keisuke Sakaguchi, Ronan~Le Bras, Chandra Bhagavatula, and Yejin Choi. 2020.
\newblock Winogrande: An adversarial winograd schema challenge at scale.
\newblock In \emph{Proceedings of the National Conference on Artificial
  Intelligence (AAAI)}.

\bibitem[{Sap et~al.(2019)Sap, Rashkin, Chen, Bras, and Choi}]{Sap2019SocialIC}
Maarten Sap, Hannah Rashkin, Derek Chen, Ronan~Le Bras, and Yejin Choi. 2019.
\newblock Social iqa: Commonsense reasoning about social interactions.
\newblock In \emph{Proceedings of the Conference on Empirical Methods for
  Natural Language Processing (EMNLP)}.

\bibitem[{Singer et~al.(1992)Singer, Halldorson, Lear, and
  Andrusiak}]{singer1992validation}
Murray Singer, Michael Halldorson, Jeffrey~C Lear, and Peter Andrusiak. 1992.
\newblock Validation of causal bridging inferences in discourse understanding.
\newblock In \emph{Journal of Memory and Language}, volume~31, pages 507--524.
  Elsevier.

\bibitem[{Spelke and Kinzler(2007)}]{spelke2007core}
Elizabeth~S Spelke and Katherine~D Kinzler. 2007.
\newblock Core knowledge.
\newblock In \emph{Developmental science}, volume~10, pages 89--96. Wiley
  Online Library.

\bibitem[{Talmor et~al.(2019{\natexlab{a}})Talmor, Herzig, Lourie, and
  Berant}]{talmor-etal-2019-commonsenseqa}
Alon Talmor, Jonathan Herzig, Nicholas Lourie, and Jonathan Berant.
  2019{\natexlab{a}}.
\newblock \href {https://doi.org/10.18653/v1/N19-1421} {{C}ommonsense{QA}: A
  question answering challenge targeting commonsense knowledge}.
\newblock In \emph{Proceedings of the North American Chapter of the Association
  for Computational Linguistics - Human Language Technologies (NAACL-HLT)},
  pages 4149--4158, Minneapolis, Minnesota. Association for Computational
  Linguistics.

\bibitem[{Talmor et~al.(2019{\natexlab{b}})Talmor, Herzig, Lourie, and
  Berant}]{talmor2019commonsenseqa}
Alon Talmor, Jonathan Herzig, Nicholas Lourie, and Jonathan Berant.
  2019{\natexlab{b}}.
\newblock \href {https://doi.org/10.18653/v1/N19-1421} {{C}ommonsense{QA}: A
  question answering challenge targeting commonsense knowledge}.
\newblock In \emph{Proceedings of the North American Chapter of the Association
  for Computational Linguistics - Human Language Technologies (NAACL-HLT)},
  pages 4149--4158, Minneapolis, Minnesota. Association for Computational
  Linguistics.

\bibitem[{Tsuchiya(2018)}]{tsuchiya2018performance}
Masatoshi Tsuchiya. 2018.
\newblock Performance impact caused by hidden bias of training data for
  recognizing textual entailment.
\newblock In \emph{International Conference on Language Resources and
  Evaluation (LREC)}.

\bibitem[{Wallace et~al.(2019)Wallace, Rodriguez, Feng, Yamada, and
  Boyd-Graber}]{Wallace2019Trick}
Eric Wallace, Pedro Rodriguez, Shi Feng, Ikuya Yamada, and Jordan Boyd-Graber.
  2019.
\newblock Trick me if you can: Human-in-the-loop generation of adversarial
  examples for question answering.
\newblock In \emph{Transactions of the Association for Computational
  Linguistics (TACL)}.

\bibitem[{Wang et~al.(2020)Wang, Liang, Jin, Wang, Zhu, and
  Zhang}]{wang-etal-2020-semeval}
Cunxiang Wang, Shuailong Liang, Yili Jin, Yilong Wang, Xiaodan Zhu, and Yue
  Zhang. 2020.
\newblock \href {https://www.aclweb.org/anthology/2020.semeval-1.39}
  {{S}em{E}val-2020 task 4: Commonsense validation and explanation}.
\newblock In \emph{Proceedings of the Fourteenth Workshop on Semantic
  Evaluation}, pages 307--321, Barcelona (online). International Committee for
  Computational Linguistics.

\bibitem[{Wieting and Kiela(2019)}]{wieting2019no}
John Wieting and Douwe Kiela. 2019.
\newblock No training required: Exploring random encoders for sentence
  classification.
\newblock In \emph{International Conference on Learning Representations
  (ICLR)}.

\bibitem[{Wolf et~al.(2020)Wolf, Debut, Sanh, Chaumond, Delangue, Moi, Cistac,
  Rault, Louf, Funtowicz, Davison, Shleifer, von Platen, Ma, Jernite, Plu, Xu,
  Scao, Gugger, Drame, Lhoest, and Rush}]{wolf-etal-2020-transformers}
Thomas Wolf, Lysandre Debut, Victor Sanh, Julien Chaumond, Clement Delangue,
  Anthony Moi, Pierric Cistac, Tim Rault, Rémi Louf, Morgan Funtowicz, Joe
  Davison, Sam Shleifer, Patrick von Platen, Clara Ma, Yacine Jernite, Julien
  Plu, Canwen Xu, Teven~Le Scao, Sylvain Gugger, Mariama Drame, Quentin Lhoest,
  and Alexander~M. Rush. 2020.
\newblock \href {https://www.aclweb.org/anthology/2020.emnlp-demos.6}
  {Transformers: State-of-the-art natural language processing}.
\newblock In \emph{Proceedings of the Conference on Empirical Methods for
  Natural Language Processing (EMNLP): System Demonstrations}, pages 38--45,
  Online. Association for Computational Linguistics.

\bibitem[{Wu et~al.(2017)Wu, Teney, Wang, Shen, Dick, and van~den
  Hengel}]{wu2017visual}
Qi~Wu, Damien Teney, Peng Wang, Chunhua Shen, Anthony Dick, and Anton van~den
  Hengel. 2017.
\newblock Visual question answering: A survey of methods and datasets.
\newblock In \emph{Computer Vision and Image Understanding}, volume 163, pages
  21--40. Elsevier.

\bibitem[{Yang et~al.(2018)Yang, Zhang, Urbanek, Feng, Miller, Szlam, Kiela,
  and Weston}]{yang2017mastering}
Zhilin Yang, Saizheng Zhang, Jack Urbanek, Will Feng, Alexander~H Miller,
  Arthur Szlam, Douwe Kiela, and Jason Weston. 2018.
\newblock Mastering the dungeon: Grounded language learning by mechanical
  turker descent.
\newblock In \emph{International Conference on Learning Representations
  (ICLR)}.

\bibitem[{Zellers et~al.(2018)Zellers, Bisk, Schwartz, and
  Choi}]{zellers-etal-2018-swag}
Rowan Zellers, Yonatan Bisk, Roy Schwartz, and Yejin Choi. 2018.
\newblock \href {https://doi.org/10.18653/v1/D18-1009} {{SWAG}: A large-scale
  adversarial dataset for grounded commonsense inference}.
\newblock In \emph{Proceedings of the Conference on Empirical Methods for
  Natural Language Processing (EMNLP)}, pages 93--104, Brussels, Belgium.
  Association for Computational Linguistics.

\bibitem[{Zellers et~al.(2019)Zellers, Holtzman, Bisk, Farhadi, and
  Choi}]{zellers-etal-2019-hellaswag}
Rowan Zellers, Ari Holtzman, Yonatan Bisk, Ali Farhadi, and Yejin Choi. 2019.
\newblock \href {https://doi.org/10.18653/v1/P19-1472} {{H}ella{S}wag: Can a
  machine really finish your sentence?}
\newblock In \emph{Proceedings of the Annual Meeting of the Association of
  Computational Linguistics (ACL)}, pages 4791--4800, Florence, Italy.
  Association for Computational Linguistics.

\bibitem[{Zhang et~al.(2018)Zhang, Liu, Liu, Gao, Duh, and
  Van~Durme}]{zhang2018record}
Sheng Zhang, Xiaodong Liu, Jingjing Liu, Jianfeng Gao, Kevin Duh, and Benjamin
  Van~Durme. 2018.
\newblock Record: Bridging the gap between human and machine commonsense
  reading comprehension.
\newblock \emph{arXiv preprint arXiv:1810.12885}.

\bibitem[{Zhang et~al.(2017)Zhang, Rudinger, Duh, and
  Van~Durme}]{zhang-etal-2017-ordinal}
Sheng Zhang, Rachel Rudinger, Kevin Duh, and Benjamin Van~Durme. 2017.
\newblock \href {https://doi.org/10.1162/tacl_a_00068} {Ordinal common-sense
  inference}.
\newblock In \emph{Transactions of the Association for Computational
  Linguistics (TACL)}, volume~5, pages 379--395.

\bibitem[{Zhou et~al.(2019)Zhou, Khashabi, Ning, and
  Roth}]{zhou-etal-2019-going}
Ben Zhou, Daniel Khashabi, Qiang Ning, and Dan Roth. 2019.
\newblock \href {https://doi.org/10.18653/v1/D19-1332} {{``}going on a
  vacation{''} takes longer than {``}going for a walk{''}: A study of temporal
  commonsense understanding}.
\newblock In \emph{Proceedings of the Conference on Empirical Methods for
  Natural Language Processing (EMNLP)}, pages 3363--3369, Hong Kong, China.
  Association for Computational Linguistics.

\end{thebibliography}
